%% file: main.tex
\documentclass[10pt,journal,compsoc]{IEEEtran}

\ifCLASSOPTIONcompsoc
  \usepackage[nocompress]{cite}
\else
  \usepackage{cite}
\fi
\ifCLASSINFOpdf
\else
\fi

\usepackage{amsmath}
\usepackage{amssymb}
\usepackage{algorithmic}
\usepackage{algorithm}
\usepackage{enumerate}
\usepackage{graphicx}
\usepackage{color}
\usepackage[dvipsnames]{xcolor}
\usepackage{tikz}
\usepackage{colortbl}

\ifCLASSOPTIONcompsoc
  \usepackage[caption=false,font=footnotesize,labelfont=sf,textfont=sf]{subfig}
\else
  \usepackage[caption=false,font=footnotesize]{subfig}
\fi

\usepackage{booktabs}
\usepackage{dblfloatfix}
\usepackage{url}
\usepackage{bm}
\usepackage{hyperref}
\usepackage{multirow,tabularx}
\newcolumntype{C}{>{\centering\arraybackslash}X}

\usepackage{amsthm}

\theoremstyle{plain}
\newtheorem{prop}{Proposition}
\theoremstyle{definition}

\definecolor{c1}{RGB}{244,152,178}
\definecolor{c2}{RGB}{251,211,210}
\definecolor{c3}{RGB}{253,238,238}
\begin{document}
%
\title{Towards the Spectral bias Alleviation \\by Normalizations in Coordinate Networks}
%
%
%
%

\author{Zhicheng Cai,
        Hao Zhu$^*$,
        Qiu Shen$^*$,
        Xinran Wang,
        Xun Cao
\IEEEcompsocitemizethanks{
\IEEEcompsocthanksitem H. Zhu and Q. Shen are the corresponding authors (zhuhao\_photo@nju.edu.cn, shenqiu@nju.edu.cn).
\IEEEcompsocthanksitem Z. Cai, H. Zhu, Q. Shen, X. Cao are with the School of Electronic Science and Engineering, Nanjing University, Nanjing, 210023, China.
\IEEEcompsocthanksitem X. Wang is with the Interdisciplinary Research Center for Future Intelligent Chips (Chip-X), Nanjing University, Suzhou, China. 
}
\thanks{Manuscript received April 19, 2005; revised August 26, 2015.}}

%
%

\markboth{Journal of \LaTeX\ Class Files,~Vol.~14, No.~8, August~2015}%
{Shell \MakeLowercase{\textit{et al.}}: Bare Advanced Demo of IEEEtran.cls for IEEE Computer Society Journals}

\IEEEtitleabstractindextext{%
\begin{abstract}
Representing signals using coordinate networks dominates the area of inverse problems recently, and is widely applied in various scientific computing tasks. Still, there exists an issue of spectral bias in coordinate networks, limiting the capacity to learn high-frequency components. This problem is caused by the pathological distribution of the neural tangent kernel's (NTK's) eigenvalues of coordinate networks. We find that, this pathological distribution could be improved using classical normalization techniques (batch normalization and layer normalization), which are commonly used in convolutional neural networks but rarely used in coordinate networks. We prove that normalization techniques greatly reduces the maximum and variance of NTK's eigenvalues while slightly modifies the mean value, considering the max eigenvalue is much larger than the most, this variance change results in a shift of eigenvalues' distribution from a lower one to a higher one, therefore the spectral bias could be alleviated (see Fig.~\ref{pic1}). Furthermore, we propose two new normalization techniques by combining these two techniques in different ways. The efficacy of these normalization techniques is substantiated by the significant improvements and new state-of-the-arts achieved by applying normalization-based coordinate networks to various tasks, including the image compression, computed tomography reconstruction, shape representation, magnetic resonance imaging, novel view synthesis and multi-view stereo reconstruction. \textit{Code:} \url{https://github.com/Aiolus-X/Norm-INR}.
\end{abstract}

}

\maketitle

\IEEEdisplaynontitleabstractindextext

%
\IEEEpeerreviewmaketitle

\ifCLASSOPTIONcompsoc
\IEEEraisesectionheading{\section{Introduction}\label{sec:introduction}}
\else
\section{Introduction}
\label{sec:introduction}
\fi

\begin{figure*}[t]
  \centering
  \includegraphics[width=\linewidth]{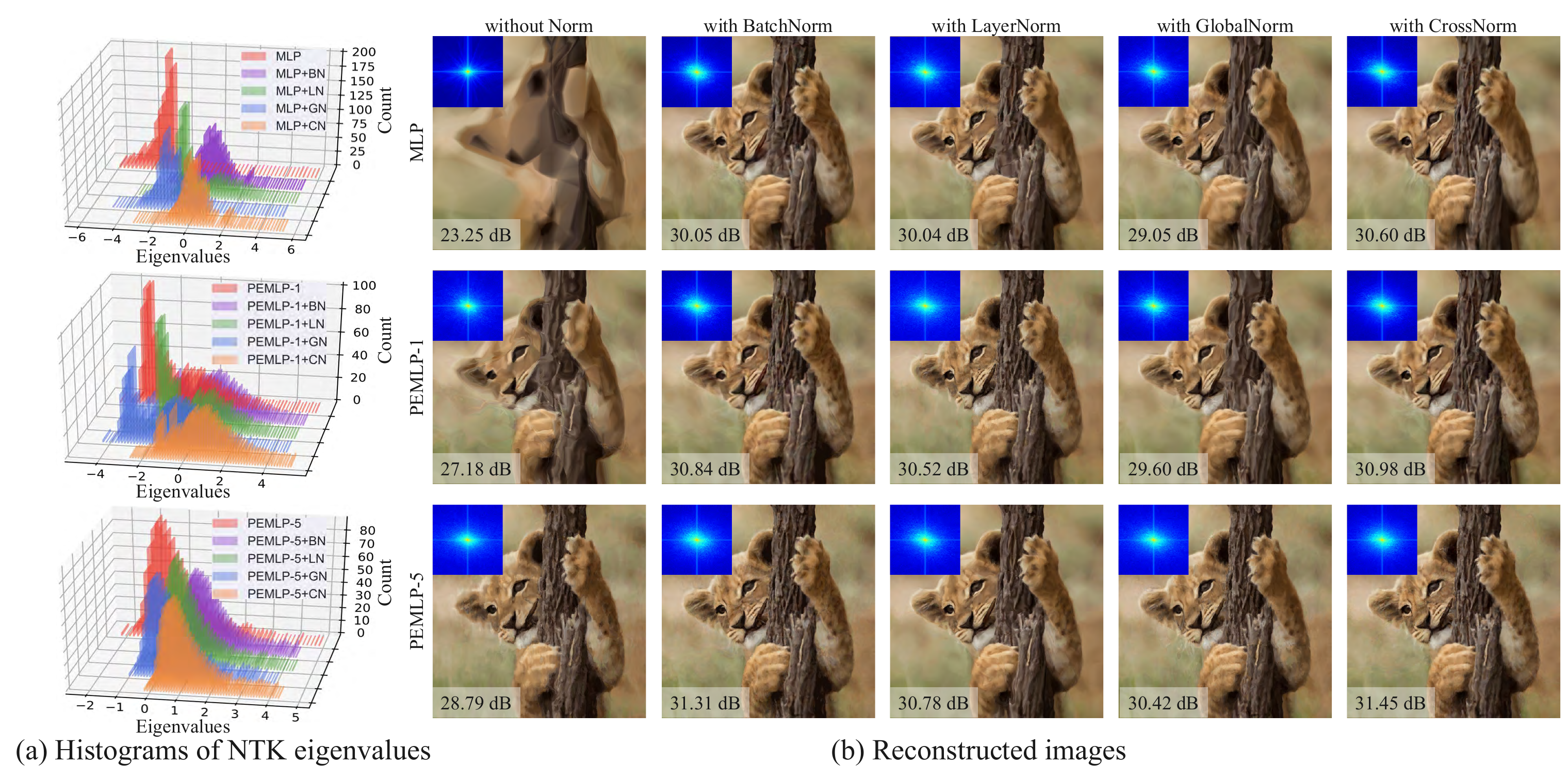}
  \caption{Normalization techniques significantly alleviate the spectral bias of coordinate networks. (a) Normalization techniques shift the NTK's eigenvalues distribution from a lower one to a higher one. From top to bottom, each row refers to the results of coordinate networks with ReLU activations, and positional encoding~\cite{tancik2020fourier} with 1 and 5 Fourier bases, respectively. Note that the values in horizontal-axis are the exponents with a base of 10. (b) The spectral bias is alleviated and better performance is achieved compared with the one without normalization (\textit{e.g.}, the texture on the lion's left paw).  }
  \label{pic1}
\end{figure*}

\IEEEPARstart{C}{oordinate} networks, which take the coordinates as inputs and output the signal attributes using multi-layer perceptron (MLP) models, have become a promising framework for solving various inverse problems. Different from the classical convolution-based networks which could only support up to 3D patterns as input~\cite{he2016deep,adnan2023untrained}, the input coordinates are organized as 1D vectors in coordinate networks, enabling the advantage of a general framework for solving inverse problems with any dimensions. Therefore, coordinate networks have been widely applied in different areas of scientific computing~\cite{karniadakis2021physics}, such as the hologram/tomography imaging in microscopy~\cite{zhu2022dnf,liu2022recovery}, 3D reconstruction and free-viewpoint roaming in computer vision/graphics~\cite{li2023neuralangelo,mildenhall2021nerf}, physical simulation in material design and hydrodynamics~\cite{chen2020physics, raissi2020hidden} and medical imaging~\cite{shen2023cardiacfield,shen2023tracking}.

Yet, due to the spectral bias~\cite{rahaman2019spectral} of ReLU-based MLP, coordinate networks prefer to learn the low-frequency components of the signal, while the high-frequency components are learned at an extremely slow convergence. Several works have been proposed to alleviate the spectral bias, such as the positional encoding~\cite{tancik2020fourier} or frequency-related activation functions~\cite{sitzmann2020implicit,ramasinghe2022beyond}. However, these explorations introduce the `frequency-specified spectral bias'~\cite{yuce2022structured}, \textit{i.e.}, only the spectrum components matching the pre-encoded frequencies could be well learned. Therefore, it is essential to encode different frequency bases as many as possible, which not only increases the complexity but also incurs the issue of suppressing lower frequency components~\cite{ramasinghe2022frequency}.

Such a spectral bias problem is related to the training dynamics of MLP. According to recent literature, the training of MLP could be viewed as kernel regression, specifically, the neural tangent kernel (NTK)~\cite{jacot2018neural,tancik2020fourier}. Following this perspective, the spectral bias is caused by the pathological distribution of NTK's eigenvalues, that most of the eigenvalues are very small, limiting the convergence speed on high-frequency components. In this paper, we find that this spectral bias could be alleviated by introducing normalization techniques, which have been widely used and studied in convolutional neural networks (CNNs) but are rarely used in the community of coordinate networks.

Different from CNNs which take 2D or 3D patterns as input, coordinate networks often take a 1D vector (\textit{i.e.}, coordinates) as inputs, consequently there are only two types of normalization techniques available, \textit{i.e.}, the batch normalization (BN) along the batch dimension and the layer normalization (LN) along the channel dimension (see Sec.~\ref{sec:norm_all} and Fig.~\ref{pic-norms} for details). We theoretically prove that both of these two methods could significantly reduce the maximum and variance of NTK's eigenvalues while applies almost no changes to the mean value. Considering the fact that the largest eigenvalue is often much larger than most ones in NTK, these maximum and variance changes could increase most of the eigenvalues. This improvement is not limited to the standard ReLU-based MLP, actually, it also works for positional encoding-based MLPs where the `frequency-specified spectral bias' could also be alleviated (as shown in Fig.~\ref{pic1}). Furthermore, we explore two potential combinations of these two normalizations in coordinate networks, namely, the global normalization (GN) and the cross normalization (CN) (see Fig.~\ref{pic-norms} for details). We find that both of these techniques could alleviate the spectral bias and the CN achieves the best performance among all four methods.

This work is extended from the preliminary exploration presented in CVPR'24~\cite{cai2024batch}. Compared with the conference version, we provide a comprehensive theoretical analysis on all possible normalization techniques in coordinate networks instead of BN only and explore two new normalization techniques (GN and CN). To verify the efficacy of these techniques, all experiments in the conference version are re-conducted and better results are achieved. Apart from this, the new proposed CN is applied to a new task, \textit{i.e.}, the multi-view stereo reconstruction, and improves state-of-the-arts.

In summary, we make the following contributions,
\begin{enumerate}
    \item We provide a comprehensive analysis on how normalization techniques improve the pathological distribution of NTK's eigenvalues, enabling the coordinate networks to learn high-frequency components effectively.
    \item We substantiate the improvements of normalization-based MLP over existing coordinate networks on two representation tasks and two inverse problems, \textit{i.e.}, image compression, shape representation, computed tomography reconstruction, and magnetic resonance imaging.
    \item We explore two novel normalization methods, \textit{i.e.}, the global normalization and cross normalization, for coordinate networks and achieve new state-of-the-arts in the areas of novel view synthesis and multi-view stereo reconstruction.
\end{enumerate}

\section{Related Work}
\subsection{Coordinate Networks}
Coordinate networks~\cite{tancik2020fourier} (also termed as \emph{implicit neural representation} or \emph{neural fields}) are gradually replacing traditional discrete representations in computer vision and graphics. Different from classical matrix-based discrete representation, coordinate networks focus on learning a neural mapping function with low-dimensional coordinates inputs and the corresponding signal values outputs, and have demonstrated the advantages of continuous querying and memory-efficient footprint in various signal representation tasks, such as images~\cite{dupont2021coin,dupont2022coin++,strumpler2022implicit}, scenes~\cite{lindell2022bacon,jiang2020local,mescheder2019occupancy,sitzmann2019scene} and videos~\cite{chen2021nerv,zhang2021implicit}. Additionally, coordinate networks could be seamlessly combined with different differentiable physical processes, opening a new way for solving various inverse problems, especially the domain-specific tasks where large-scale labelled datasets are unavailable, such as the shape representation~\cite{park2019deepsdf,michalkiewicz2019implicit,chen2019learning,genova2019learning,chabra2020deep}, computed tomography reconstruction~\cite{saragadam2023wire,molaei2023implicit,gupta2022neural,shen2022nerp,reed2021dynamic} and inverse rendering for novel view synthesis~\cite{mildenhall2021nerf,niemeyer2020differentiable,pumarola2021d,rebain2021derf,zhu2023pyramid}.

\subsection{Overcoming the Spectral Bias}
MLPs with conventional activations such as ReLU encounter the issue of \emph{spectral bias}~\cite{basri2020frequency,rahaman2019spectral,tancik2020fourier}, which restricts their ability to capture the high-frequency components present in visual signals. To overcome this limitation, two types of methods are raised in literature, namely, the function expansion and the signal re-organization. The first type of methods treats the coordinate networks as a function expansion process, \textit{e.g.}, the Fourier, Taylor, Wavelet and Gaussian expansions, as a result, more pre-encoded bases could improve the expressive power. Following this idea, various bases have been introduced to encode the input coordinates or activate the hidden neurons, such as the Fourier/polynomial encodings~\cite{tancik2020fourier,landgraf2022pins,Singh_2023_CVPR}, periodic/non-periodic/wavelet activation functions~\cite{sitzmann2020implicit,ramasinghe2022beyond,saragadam2023wire,liu2024finer}, and compositing multiple filters~\cite{fathony2020multiplicative,lindell2022bacon}. However, these methods are sensitive to the training configurations. 

Another type of methods focuses on mapping the input complex signal to another one which is composed of low-frequency components~\cite{takikawa2021neural,chabra2020deep,jiang2020local,muller2022instant,kang2022pixel,zhu2023disorder}, thus the original signal could be well learned. This mapping function is often implemented by introducing learnable hash tables between the input coordinate and the subsequent neural network, such as the single scale full-resolution hash table used in DINER~\cite{zhu2023disorder}, multi-scale pyramid hash tables in InstantNGP~\cite{muller2022instant} and multiple shifting hash tables in PIXEL~\cite{kang2022pixel}. These methods achieve high performance for representing complex signals at the cost of losing ability for interpolation, often requiring additional regularizations~\cite{zhu2023rhino}. 

Different from previous methods, this paper raises a novel way to alleviate the spectral bias by introducing normalization methodology to the coordinate networks area.

\subsection{Normalization for Deep Learning}
Normalization~\cite{ioffe2015batch,ba2016layer,ulyanov2016instance,miyato2018spectral,salimans2016weight,wu2018group} has been an indispensable methodology for deep learning~\cite{huang2023normalization,cai2023falconnet,liu2022convnet,liu2022swin,vaswani2017attention}. 
As a milestone technique, batch normalization (BN)~\cite{ioffe2015batch} is raised to solve the issue of \emph{internal covariate shift}, which improves training stability, optimization efficiency and generalization ability~\cite{morcos2018importance,kohler2018towards}.
It has been a fundamental component for modern visual models and successfully applied to a wide range of computer vision tasks~\cite{tan2019efficientnet,tan2021efficientnetv2,chen2023run}. 
Furthermore, many variants of BN are proposed, such as Conditional BN~\cite{de2017modulating} and Decorrelated BN~\cite{huang2018decorrelated}.
In the realm of theoretical analysis,
\cite{im2016empirical} highlights that BN tends to reduce the dependency on weight initialization in optimization trajectories.
\cite{santurkar2018does,dukler2020optimization} prove that BN improves optimization by mitigating the pathological curvature and smoothing the loss landscape. 
\cite{bjorck2018understanding,luo2018towards} prove that BN permits using a larger learning rate according to the gradient dynamics and random matrix theory.

Layer normalization (LN)~\cite{ba2016layer} is another representative normalization technique, unlike BN which conducts the normalization along the batch dimension, LN normalizes the feature maps in the channel dimension, guaranteeing the invariance under weights and data transformations. 
LN has been a critical component especially for transformers~\cite{liu2022swin,vaswani2017attention} and large language models~\cite{touvron2023llama,achiam2023gpt}, successfully applied to various series predicting~\cite{gu2023mamba,wu2023interpretable} and natural language processing tasks~\cite{vaswani2017attention,devlin2018bert,touvron2023llama,achiam2023gpt}. 
Additionally, many LN variations are proposed, such as Dynamic LN~\cite{kim2017dynamic} and Root-mean-square LN~\cite{zhang2019root}.
Theoretically, \cite{ba2016layer} illustrates that LN  stabilizes the convergence through learning the magnitude of incoming weights. Besides, \cite{lyu2022understanding} analyses that LN benefits the generalization though reducing the sharpness of loss surface.

This paper provides a comprehensive theoretical analysis and empirical evidence demonstrating that normalization can effectively overcome the spectral issues, thus enhancing the representational capacity of coordinate networks significantly.
Furthermore, this paper proposes two novel normalization methods, namely, the global normalization and cross normalization, especially for coordinate networks, achieving new state-of-the-arts in various representation tasks and inverse problems.

\section{Background}
To lay the foundation for our theoretical analysis, we give the background of coordinate networks including the formulation and the neural tangent kernel. In order to facilitate a better understanding of the derivations in the following sections, Tab.~\ref{tab:notations} provides the notations used throughout the paper. 

\begin{table}[t]
\footnotesize
\begin{center}
\caption{Notations}
\label{tab:notations}
\begin{tabular}{ll}
\toprule
Notation & Definition\\ 
\midrule
$T$           & total number of training samples\\
$L$           & number of hidden layers of MLP\\
$N^0$         & length of each input coordinates vector\\
$N,N^l$         & width of the $l\!-\!th$ hidden layer, $l\in[1,L\!-\!1]$\\
$C,N^L$      & dimensions of target signal\\
$|\theta|$    & total number of parameters of MLP\\
$f(\mathbf{X};\theta),\mathbf{H}^{L}$ & final output of MLP, $\in\mathbb{R}^{C\times T}$\\
$f^{BN}(\mathbf{X};\theta)$&final output of MLP with BN, $\in\mathbb{R}^{C\times T}$\\
$f^{LN}(\mathbf{X};\theta)$&final output of MLP with LN, $\in\mathbb{R}^{C\times T}$\\
$\mathbf{M}^{BN}$ & mean value matrix of BN, $\in\mathbb{R}^{C\times T}$\\
$\mathbf{V}^{BN}$ & variance value matrix of BN, $\in\mathbb{R}^{C\times T}$\\
$\mathbf{M}^{LN}$ & mean value matrix of LN, $\in\mathbb{R}^{C\times T}$\\
$\mathbf{V}^{LN}$ & variance value matrix of LN, $\in\mathbb{R}^{C\times T}$\\
$\mathbf{K}$ & NTK of standard MLP, $\in\mathbb{R}^{C\times T\times T\times C}$\\
$\mathbf{K}^{BN}$ & NTK of MLP with BN, $\in\mathbb{R}^{C\times T\times T\times C}$\\
$\mathbf{K}^{LN}$ & NTK of MLP with LN, $\in\mathbb{R}^{C\times T\times T\times C}$\\
$\nabla_{\theta}f(\mathbf{X};\theta),\nabla_{\theta}{\mathbf{H}^{L}}$ & deviation of MLP output to $\theta$, $\in\mathbb{R}^{|\theta| \times T\times C}$\\
$\nabla_{\theta}f^{BN}(\mathbf{X};\theta)$& deviation of BN-MLP output to $\theta$, $\in\mathbb{R}^{|\theta| \times T\times C}$\\
$\nabla_{\theta}f^{LN}(\mathbf{X};\theta)$& deviation of LN-MLP output to $\theta$, $\in\mathbb{R}^{|\theta| \times T\times C}$\\
$p_t^l, q_t^l, p^l_{st}, q^l_{st}$& order parameters of mean field theory\\
$\kappa_1, \kappa_2$& constants corresponding to order parameters\\
$\mathbf{A}^{BN}$ & projector of BN variance division, $\in\mathbb{R}^{C\times T\times T\times C}$\\
$\mathbf{G}^{BN}$ & projector of BN mean subtraction, $\in\mathbb{R}^{C\times T\times T\times C}$\\
$\mathbf{A}^{LN}$ & projector of LN variance division, $\in\mathbb{R}^{C\times T\times T\times C}$\\
$\mathbf{G}^{LN}$ & projector of LN mean subtraction, $\in\mathbb{R}^{C\times T\times T\times C}$\\
\bottomrule		
\end{tabular}
\end{center}
\end{table}


\subsection{Coordinate Networks}
Given a signal $\{\vec{x}_i,\vec{y}_i\}_{i=1}^{T}$ with $T$ samples, coordinate networks learn a function that maps input coordinates to the output corresponding signal values. It is often parameterized by a multi-layer perceptron (MLP) network consisting of one input layer with $N^0$ neurons and $L$ hidden layers (including one output layer) with $N^l$ neurons per hidden layer ($l=1,2,...,L$).  Without a loss of generality, we focus on a multi-dimensional signal with $C$ attributes (\textit{i.e.}, $N^{L}=C$), additionally all other hidden layers have same $N$ neurons (\textit{i.e.}, $N^1=...=N^{L-1}=N$). Thus we can describe the network $f(\mathbf{X};\theta)$ as,
\begin{equation}
\begin{aligned}
    &\mathbf{H}^{0}=\mathbf{X}=[\vec{x}_1,\vec{x}_2,...,\vec{x}_T] \\
    &\mathbf{H}^{l}=\phi(\mathbf{W}^{l}\mathbf{H}^{l-1} + \vec{b}^{\:l})\\
    &f(\mathbf{X};\theta)=\mathbf{H}^{L}=\mathbf{W}^{L}\mathbf{H}^{L-1}+\vec{b}^{L}
\end{aligned}
\label{mlp}
\end{equation}
where $\mathbf{X}\in \mathbb{R}^{N^{0}\times T}$ refers to the matrix of all coordinates in the training dataset, respectively. $\theta = \{\mathbf{W}^{l},\vec{b}^{\:l} \ |\  l=1,...,L\}$ is the network parameters in $f(\mathbf{X};\theta)$. $\mathbf{W}^{l}\in \mathbb{R}^{N^{l}\times N^{l-1}}$ and $\vec{b}^{\:l}\in \mathbb{R}^{N^{l}}$ are the weight matrix and bias vector of the $l$-th layer, respectively, and are randomly initiated by Gaussian distribution with mean value of $0$ and variance of $\sqrt{\sigma_{w}^2/N^{l-1}}$ . 
$\phi(\cdot)$ is the activation function, and $\mathbf{H}^{l}\in \mathbb{R}^{N^{l}\times T}$ is the output matrix of the $l$-th layer.

\subsection{Neural Tangent Kernel and Spectral Bias}
\label{sec:ntk_bias}
Neural tangent kernel (NTK)~\cite{jacot2018neural,tancik2020fourier}, which approximates the training of neural network as a linear kernel regression, has become a popular lens for monitoring the dynamic behaviors and convergence of a neural network. Given a neural network $f(\mathbf{X};\theta)$, its NTK is defined as,
\begin{equation}
\begin{aligned}
\mathbf{K}&=(\nabla_{\theta}f(\mathbf{X};\theta))^{\top}\nabla_{\theta}f(\mathbf{X};\theta)\\
&=(\nabla_{\theta}\mathbf{H}^{L})^{\top}\nabla_{\theta}\mathbf{H}^{L}.
\end{aligned}
\label{eqn:ntk}
\end{equation}

When the network $f(\mathbf{X};\theta)$ is trained following an $L_2$ loss function, SGD optimizer and a learning rate $\eta$, the network output after $\alpha$ training iterations can be approximated by the NTK as~\cite{tancik2020fourier,lee2019wide}:
\begin{equation}
    \begin{split}
    \mathbf{Y}^{(n)} \approx \left(\mathbf{I}-e^{-\eta\mathbf{K}\alpha}\right)\mathbf{Y},
    \label{ntk-output}
    \end{split}
\end{equation}
where $\mathbf{I}$ is the identity matrix, $\mathbf{Y}$ refers to signal values of all points in training dataset, namely, $\mathbf{Y}=[\vec{y}_1,\vec{y}_2,...,\vec{y}_T]$. Since the NTK matrix $\mathbf{K}$ is a positive semi-definite matrix, it could be eigendecomposed using SVD as $\mathbf{K}=\mathbf{Q}\mathbf{\Lambda}\mathbf{Q}^{\top}$, where $\mathbf{Q}$ is a orthogonal matrix composed of the eigenvectors and $\mathbf{\Lambda}$ is a diagonal matrix full of the eigenvalues $\lambda_i \geq 0$ of $\mathbf{K}$. Thus we have $e^{-\eta\mathbf{K}\alpha}=\mathbf{Q}e^{-\eta\mathbf{\Lambda}\alpha}\mathbf{Q}^{\top}$, and the training error could be modelled as,
\begin{equation}
\begin{aligned}
    &|\mathbf{Y}^{(n)}-\mathbf{Y}| \approx  e^{-\eta\mathbf{K}\alpha}\mathbf{Y} =\mathbf{Q}e^{-\eta\mathbf{\Lambda}\alpha}\mathbf{Q}^{\top}\mathbf{Y} \\
    \Rightarrow & \mathbf{Q}^{\top}|\mathbf{Y}^{(n)}-\mathbf{Y}|\approx e^{-\eta\mathbf{\Lambda}\alpha}\mathbf{Q}^{\top}\mathbf{Y}.
\label{spec-bias}
\end{aligned}
\end{equation}
It could be noticed that the training error is determined by the eigenvalues in $\mathbf{\Lambda}$. The network $f(\mathbf{X};\theta)$ could learn the components with a large eigenvalue rapidly, and has a slow convergence to small eigenvalues which often refer to the high frequency components of the signal to be learned~\cite{tancik2020fourier}. This phenomenon is termed as `spectral bias'.

\subsection{NTK Elements and Eigenvalues}
\subsubsection{Calculation of NTK Elements}
According to the mean field theory~\cite{karakida2019universal,karakida2019normalization}, each element $\mathbf{K}_{i,j,s,t}=(\nabla_{\theta}\mathbf{H}^L_{i,t})^{\top}\nabla_{\theta}\mathbf{H}^L_{j,s}$ in the $s,t$-th pixel ($t,s=1,...,T$) of the $i,j$-th block ($i,j=1,...,C$) of the $C\times T\times T\times C$ coordinate network NTK $\mathbf{K}=(\nabla_{\theta}\mathbf{H}^L)^{\top}\nabla_{\theta}\mathbf{H}^L$ can be approximately calculated and re-written in the explicit numerical form as:

\begin{equation}
    \begin{split}
&\mathbf{K}_{i,j,s,t}= \begin{cases} 
    N\kappa_1 + \mathcal{O}(\sqrt{N}),     &  for\ \  i=j,     s=t    \\
    N\kappa_2 + \mathcal{O}(\sqrt{N}),     &  for\ \  i=j,     s\neq t\\
    \mathcal{O}(\sqrt{N}),                 &  for\ \  i\neq j, s=t    \\
    \mathcal{O}(\sqrt{N}),                 &  for\ \  i\neq j, s\neq t
    \end{cases}\\
&\kappa_1 = \sum^L_{l=1} p_t^{l-1} q_t^{l}, \ \ \ \ \kappa_2 = \sum^L_{l=1} p^{l-1}_{st} q^{l}_{st}.
\end{split}
    \label{eq-1}
\end{equation}
$\kappa_1$ and $\kappa_2$ are two constant values corresponding to the order parameters of the mean field theory $p_t^l, q_t^l, p^l_{st}, q^l_{st}$~\cite{karakida2019universal}. 
In brief, $p^l_{t},p^l_{st}$ are the order parameters for the forward signal propagation, utilized to explain the depth to which signals can be sufficiently propagated from the perspective of order-to-chaos phase transition~\cite{jacot2019order}. Similarly, $q^l_{t},q^l_{st}$ are the order parameters for the backward gradient propagation.

\subsubsection{Statistical Characteristics of NTK's Eigenvalues}
Because it is difficult and infeasible to calculate the detailed numerical values of NTK's eigenvalues of a random initialized MLP network, we focus on analysing its statistic characteristics, \textit{i.e.}, mean value $m_{\lambda}$, variance $v_{\lambda}$ and the maximum value $\lambda_{max}$. According to the matrix theory~\cite{franklin2012matrix}, these values could be calculated or estimated by,
\begin{equation}
    \begin{split}
    &m_\lambda = \frac{1}{CT}Trace(\mathbf{K}) = \frac{1}{CT}\sum^C_{i=1}\sum^T_{t=1}\mathbf{K}_{i,i,t,t}\\
    &s_\lambda \!=\! \frac{1}{CT}Trace(\mathbf{K}\mathbf{K}^{\top}) \!=\! \frac{1}{CT}\sum^C_{i=1}\sum^T_{s=1}\sum^C_{j=1}\sum^T_{t=1}\mathbf{K}_{i,j,s,t}^2\\
    &v_\lambda = s_\lambda-m_\lambda^2 \\
    & \frac{s_\lambda}{m_\lambda} = \frac{\sum^{CT}_{i=1}\lambda_i^2}{\sum^{CT}_{i=1}\lambda_i} \leq \lambda_{max} \leq   \sqrt{\sum^{CT}_{i=1}\lambda_i^2}=\sqrt{CTs_\lambda},
    \end{split}
    \label{eq-7}
\end{equation}
where $s_{\lambda}$ is the second momentum. According to the explicit element values of the NTK in the above section, these statistical values of standard MLP could be obtained directly,
\begin{equation}
\begin{aligned}
&\text{Standard MLP:}\\
&\qquad m_\lambda = \frac{NCT\kappa_1+\mathcal{O}(\sqrt{N})CT}{CT} \sim \mathcal{O}(N\!+\!\sqrt{N})\\
&\qquad s_\lambda \sim \mathcal{O}\left(T(N\!+\!\sqrt{N})^2\!+\!(CT\!-\!T)N\right) \\
&\qquad v_\lambda \sim \mathcal{O}\left((N\!+\!2\sqrt{N}\!+\!C)TN\right) \\
&\qquad \mathcal{O}(NT\!+\!\sqrt{N}T\!+\!CT) \leq \lambda_{max} \leq  \mathcal{O}(N\sqrt{C}T\!+\!\sqrt{N}CT).\\
\end{aligned}
\label{eqn:ntk_cmp_mlp}
\end{equation}
\noindent \textbf{Pathological distribution of NTK eigenvalues causes spectral bias.} 
As shown in Eqn.~\ref{eqn:ntk_cmp_mlp}, the NTK eigenvalue distribution of MLP has a relatively small mean value with the scale of $\mathcal{O}(N\!+\!\sqrt{N})$, while the largest eigenvalue is significantly enormous with an upper bound of $\mathcal{O}(N\sqrt{C}T\!+\!\sqrt{N}CT)$. Additionally, the variance is also extremely large with the scale of $\mathcal{O}\left((N\!+\!2\sqrt{N}\!+\!C)TN\right)$, thus most of the other eigenvalues are close to zero, resulting in the issue of pathological distribution of eigenvalues. The eigenvalue distribution of a standard MLP in Fig.~\ref{pic1}\textcolor{red}{a} (red histogram) also verifies the phenomenon of pathological distribution.
As stated above, the corresponding frequencies with small eigenvalues are consequently learned at a extremely slow rate, incurring the issue of spectral bias.

\begin{figure}[t]
  \centering
  \includegraphics[width=\linewidth]{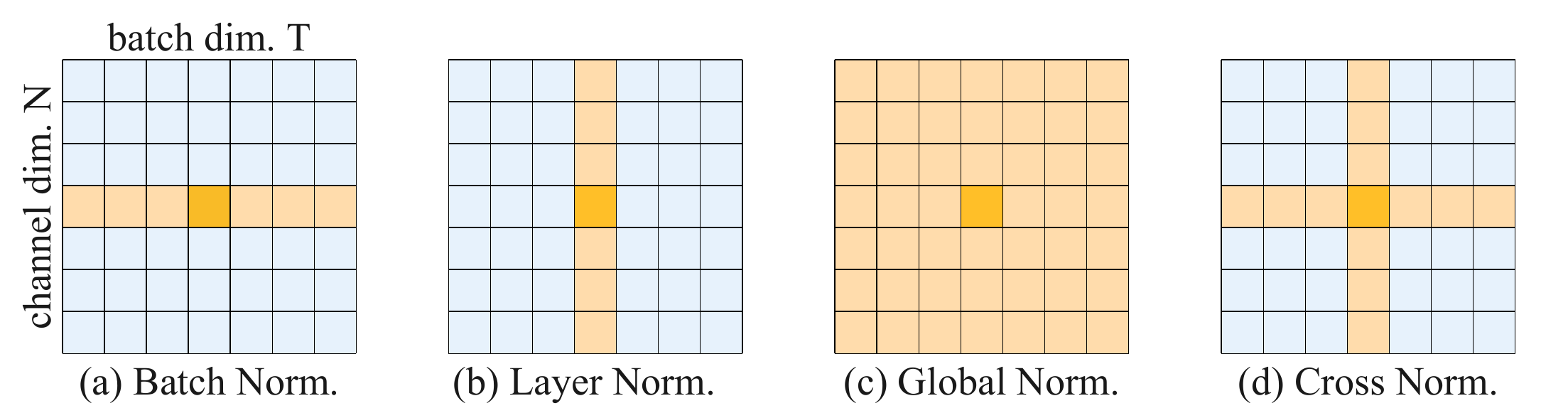}
  \caption{Schematic diagrams of different normalization techniques. }
  \label{pic-norms}
\end{figure}
\section{Normalization on Coordinate Networks}
\label{sec:norm_all}
Normalization techniques have been well explored in convolutional neural networks (CNNs) and various techniques have been proposed~\cite{huang2023normalization} but are rarely adopted and discussed in the community of coordinate networks. Different from CNNs which take 2D or 3D patterns, coordinates networks take 1D coordinate vector as input, consequently some normalization techniques become unavailable, such as the group normalization and instance normalization. Fig.~\ref{pic-norms} visualizes the structure of the input of coordinate networks. It is observed that there are two dimensions, \textit{i.e.}, the batch- and the channels-dimension. In this section, we first discuss the normalization along the batch and channels in detail, \textit{i.e.}, the BN and the LN. Then two new normalization techniques, \textit{i.e,} the global normalization (GN) and cross normalization (CN) are proposed by summarizing the derivations of BN and LN.

\subsection{Batch normalization}
As the name says, batch normalization aims at normalizing the network output at different layers along the batch dimension. Without a loss of generality, we starts from the simplest case, \textit{i.e.}, applying BN to the last layer of $f(\mathbf{X};\theta)$, thus the element $f(\mathbf{X};\theta)_{i,t}$ on the $i$-th row ($i=1,2,...,N$) and $t$-th column ($t=1,2,...,T$) of $f(\mathbf{X};\theta)$ becomes, 
\begin{equation}
\begin{aligned}
    &f^{BN}(\mathbf{X};\theta)_{i,t}=\frac{{\mathbf{H}_{i,t}}^{L}-{\mu_i}}{\sigma_i}\gamma_i + \beta_i\\
    &\mu_i \!=\! \frac{1}{T}\sum^T_{n=1}\mathbf{H}^{L}_{i,n},\ \ \sigma_i\!=\!\sqrt{\frac{1}{T}\sum^T_{n=1}(\mathbf{H}^{L}_{i,n})^2\!-\!(\frac{1}{T}\sum^T_{n=1}\mathbf{H}^{L}_{i,n})^2},
\end{aligned}
\label{eqn:bn}
\end{equation}
where 
the scale and shift parameters $\gamma_i$ and $\beta_i$ are initialized as $1$ and $0$, respectively, as a common practice~\cite{yang2019mean,jacot2019order}. In addition, we denote the mean value matrix $\mathbf{M}^{BN}\in \mathbb{R}^{C\times T}$ consisting of $[\mu_1,\mu_2,...,\mu_C]$, and the variance value matrix $\mathbf{V}^{BN}\in \mathbb{R}^{C\times T}$ consisting of $[\sigma_1,\sigma_2,...,\sigma_C]$, to be specific, $\mathbf{M}^{BN}_{i,t}=\mu_i$ and $\mathbf{V}^{BN}_{i,t}=\sigma_i$.

In this case, the NTK changes and could be derived by substituting the Eqn.~\ref{eqn:bn} into Eqn.~\ref{eqn:ntk}. And the modified NTK $\mathbf{K}^{BN}$ could be directly calculated by the chain rule of differentiation as,
\begin{equation}
\begin{aligned}
    \mathbf{K}^{BN}&=(\nabla_{\theta}f^{BN}(\mathbf{X};\theta))^{\top}\nabla_{\theta}f^{BN}(\mathbf{X};\theta)\\
    &=\frac{(\nabla_{\theta}\mathbf{H})^{\top}\nabla_{\theta}\mathbf{H}}{({\mathbf{V}^{BN}})^2} - \frac{\mathbf{H}^{\top}\mathbf{H}(\nabla_{\theta}\mathbf{H})^{\top}\nabla_{\theta}\mathbf{H}}{T{(\mathbf{V}^{BN})}^4},
\end{aligned}
\label{eqn:ntk_bn}
\end{equation}
where $\mathbf{H}=\mathbf{H}^{L}-{\mathbf{M}^{BN}}$ is introduced for the sake of derivation. Please refer the supplemental material for details of the derivation.

\subsubsection{Eigenvalues of BN-based MLP}

Furthermore, we can equivalently rewrite the NTK with batch normalization (\textit{i.e.}, Eqn.~\ref{eqn:ntk_bn}) as,
\begin{equation}
    \begin{split}
    \mathbf{K}^{BN}
     = \mathbf{A}^{BN}(\nabla_{\theta}\mathbf{H}^L-\nabla_{\theta}{\mathbf{M}^{BN}})^{\top}(\nabla_{\theta}\mathbf{H}^L-\nabla_{\theta}{\mathbf{M}^{BN}}),
    \end{split}
    \label{eq-2}
\end{equation}
where $\mathbf{A}^{BN}$ is a $C\times T\times T\times C$ mapping tensor corresponding to the variance division, specifically:
\begin{equation}
    \begin{split}
\mathbf{A}^{BN}_{i,j,s,t}\!=\! \begin{cases} 
    \frac{1}{\sigma_i\sigma_j}(1\!-\!\frac{1}{T}\frac{\sum^T_n \textbf{H}_{i,n}\textbf{H}_{j,n} }{\sigma_i\sigma_j}),&for\ \  i=j,     s=t    \\
    \frac{1}{\sigma_i\sigma_j}(-\frac{1}{T}\frac{\sum^T_n \textbf{H}_{i,n}\textbf{H}_{j,n} }{\sigma_i\sigma_j}),&for\ \  i=j,     s\neq t\\
    0,             &for\ \  i\neq j, s=t    \\
    0,             &for\ \  i\neq j, s\neq t
    \end{cases}.
    \end{split}
    \label{eq-3}
\end{equation}
Here we further define the $C\times T\times T\times C$ projector tensor $\mathbf{G}^{BN}$ corresponding to the mean subtraction:
\begin{equation}
    \begin{split}
\mathbf{G}^{BN}_{i,j,s,t}= \begin{cases} 
    \frac{T-1}{T}, &  for\ \  i=j,     s=t    \\
    -\frac{1}{T},  &  for\ \  i=j,     s\neq t\\
    0,             &  for\ \  i\neq j, s=t    \\
    0,             &  for\ \  i\neq j, s\neq t
    \end{cases}.
    \end{split}
    \label{eq-4}
\end{equation}
$\mathbf{G}^{BN}$ satisfies $(\mathbf{G}^{BN})^{\top}\mathbf{G}^{BN}=\mathbf{G}^{BN}$ and $\nabla_{\theta}\mathbf{H}^L\mathbf{G}^{BN}=(\nabla_{\theta}\mathbf{H}^L-\nabla_{\theta}{\mathbf{M}^{BN}})$.
Thus Eqn.~\ref{eq-2} becomes:
\begin{equation}
    \begin{split}
    \mathbf{K}^{BN} &= \mathbf{A}^{BN} (\mathbf{G}^{BN})^{\top} (\nabla_{\theta}\mathbf{H}^L)^{\top}\nabla_{\theta}\mathbf{H}^L \mathbf{G}^{BN}\\
    &= \mathbf{A}^{BN} (\mathbf{G}^{BN})^{\top} \mathbf{K} \mathbf{G}^{BN}.\\
    \end{split}
    \label{eq-5}
\end{equation}
Thus we can numerically calculate each element in $\mathbf{K}^{BN}$ as\footnote{Please refer the supplemental material for details of derivation.}:
\begin{equation}
    \begin{split}
\mathbf{K}^{BN}_{i,j,s,t}\approx \begin{cases} 
    N\frac{\kappa_1-\kappa_2}{\sigma_i\sigma_j} + \mathcal{O}(\sqrt{N}),     &  for\ \  i=j,     s=t    \\
    \frac{N}{T}\frac{\kappa_2-\kappa_1}{\sigma_i\sigma_j} + \mathcal{O}(\sqrt{N}),     &  for\ \  i=j,     s\neq t\\
    \mathcal{O}(\sqrt{N}),                          &  for\ \  i\neq j, s=t    \\
    \mathcal{O}(\sqrt{N}),                          &  for\ \  i\neq j, s\neq t
    \end{cases}.
    \end{split}
    \label{eq-6}
\end{equation}

By substituting the Eqn.~\ref{eq-6} into the Eqn.~\ref{eq-7}, the statistical characteristics of NTK's eigenvalues in BN-based MLP could be obtained,
\begin{equation}
\begin{aligned}
&\text{BN-based MLP:}\\
&\qquad m_\lambda = \frac{NT\sum_{i=1}^C\!\frac{(\kappa_1\!-\!\kappa_2)}{\sigma_i^2}\!+\!\mathcal{O}(\sqrt{N})CT}{CT} \sim \mathcal{O}(N\!+\!\sqrt{N})\\
&\qquad s_\lambda \sim \mathcal{O}\left((N\!+\!\sqrt{N})^2\!+\!(CT\!-\!1)N\right) \\
&\qquad v_\lambda \sim \mathcal{O}(CTN) \\
&\qquad \mathcal{O}(N\!+\!\sqrt{N}\!+\!CT) \leq \lambda_{max} \leq   \mathcal{O}(N\sqrt{CT}\!+\!\sqrt{N}CT).
\end{aligned}
\label{eqn:ntk_cmp_bn}
\end{equation}

\vspace{0.5em}
\noindent \textbf{Batch normalization alleviates the spectral bias.} 
Comparing these NTK eigenvalues statistical characteristics before and after applying batch normalization to the MLP (Eqn.~\ref{eqn:ntk_cmp_bn} \textit{vs}. Eqn.~\ref{eqn:ntk_cmp_mlp}), it is observed that the mean values are slighted changed and maintain the same scale of $\mathcal{O}(N\!+\!\sqrt{N})$. However, the variances are significantly reduced from $\mathcal{O}\left((N\!+\!2\sqrt{N}\!+\!C)TN\right)$ to $\mathcal{O}\left(CTN\right)$, considering the largest eigenvalue is also largely reduced by the scale of $\mathcal{O}(\sqrt{T})$, this variance change indicates a shift of eigenvalues' distribution from a lower one to a higher one (namely, most of the eigenvalues are significantly enlarged, as shown in Fig.~\ref{pic1}\textcolor{red}{a}). As a result, the issue of spectral bias could be alleviated. Although the analysis mentioned above is built upon the assumption of applying BN to the last layer, it could be expanded to a general case by modeling deep BN-based coordinate networks as a stacking process layer by layer. 

\subsection{Layer Normalization}
We have theoretically and experimentally validated that adding normalization in the batch dimension alleviates the spectral bias. In the subsection, we further explore the effectiveness of employing normalization in the channel dimension, i.e., the Layer Normalization (LN)~\cite{ba2016layer}.
Similarly, we apply LN to the last layer of the coordinate network $f(\mathbf{X};\theta)$, thus the element $f(\mathbf{X};\theta)_{i,t}$ accordingly becomes, 
\begin{equation}
\begin{aligned}
    &f^{LN}(\mathbf{X};\theta)_{i,t}=\frac{{\mathbf{H}_{i,t}}^{L}-{\mu_t}}{\sigma_t}\gamma_i + \beta_i\\
    &\mu_t \!=\! \frac{1}{C}\sum^C_{m=1}\mathbf{H}^{L}_{m,t},\ \ \sigma_t\!=\!\sqrt{\frac{1}{C}\sum^C_{m=1}(\mathbf{H}^{L}_{m,t})^2\!-\!(\frac{1}{C}\sum^C_{m=1}\mathbf{H}^{L}_{m,t})^2},
\end{aligned}
\label{eqn:ln}
\end{equation}
where the scale and shift parameters $\gamma_i$ and $\beta_i$ are also initialized as $1$ and $0$. 
In this case, we denote the mean value matrix $\mathbf{M}^{LN}\in \mathbb{R}^{C\times T}$ consisting of $[\mu_1,\mu_2,...,\mu_T]$, and the variance value matrix $\mathbf{V}^{LN}\in \mathbb{R}^{C\times T}$ consisting of $[\sigma_1,\sigma_2,...,\sigma_T]$, to be specific, $\mathbf{M}^{LN}_{i,t}=\mu_t$ and $\mathbf{V}^{LN}_{i,t}=\sigma_t$.

Correspondingly, the modified NTK $\mathbf{K}^{LN}$ could be directly calculated by the chain rule of differentiation as,
\begin{equation}
\begin{aligned}
    \mathbf{K}^{LN}&=\left(\nabla_{\theta}f^{LN}(\mathbf{X};\theta)\right)^{\top}\nabla_{\theta}f^{LN}(\mathbf{X};\theta)\\
    &=\left(\frac{\nabla_{\theta}\textbf{H}^{\top}}{\mathbf{V}^{LN}}\!-\!\frac{\textbf{H}^{\top}\textbf{H}\nabla_{\theta}\textbf{H}^{\top}}{C(\mathbf{V}^{LN})^3}\right)\left(\frac{\nabla_{\theta}\textbf{H}}{\mathbf{V}^{LN}}\!-\!\frac{\nabla_{\theta}\textbf{H}\textbf{H}^{\top}\textbf{H}}{C(\mathbf{V}^{LN})^3}\right).
\end{aligned}
\label{eqn:ntk_ln}
\end{equation}
still, $\mathbf{H}=\mathbf{H}^{L}-{\mathbf{M}^{LN}}$ is introduced for the sake of derivation. 

\subsubsection{Eigenvalues of LN-based MLP}
Similarly, we define the $C\times T\times T\times C$ mapping tensor $\mathbf{A}^{LN}$ corresponding to the variance division
\begin{equation}
    \begin{split}
\mathbf{A}^{LN}_{i,j,s,t}\!=\! \begin{cases} 
    \frac{1}{\sigma_t\sigma_s}(1\!-\!\frac{\sum^C_m \textbf{H}_{m,s}\textbf{H}_{m,t}}{C\sigma_t\sigma_s}),  &for\ \  i=j,     s=t    \\
    0,             &for\ \  i=j,  s\neq t\\
    \frac{1}{\sigma_t\sigma_s}(\frac{1}{C}\!-\!\frac{\sum^C_m \textbf{H}_{m,s}\textbf{H}_{m,t}}{C\sigma_t\sigma_s}),      &for\ \  i\neq j, s=t    \\
    0,             &for\ \  i\neq j, s\neq t
    \end{cases}.
    \end{split}
    \label{eqn:a_ln}
\end{equation}
Additionally, we define the $C\times T\times T\times C$ projector tensor $\mathbf{G}^{LN}$ corresponding to the mean subtraction:
\begin{equation}
    \begin{split}
\mathbf{G}^{LN}_{i,j,s,t}= \begin{cases} 
    \frac{C-1}{C}, &  for\ \  i=j,     s=t    \\
    0,             &  for\ \  i=j,     s\neq t\\
    -\frac{1}{C},  &  for\ \  i\neq j, s=t    \\
    0,             &  for\ \  i\neq j, s\neq t
    \end{cases}.
    \end{split}
    \label{eqn:g_ln}
\end{equation}
Still, $\mathbf{G}^{LN}$ satisfies $(\mathbf{G}^{LN})^{\top}\mathbf{G}^{LN}\!=\!\mathbf{G}^{LN}$ and $\nabla_{\theta}\mathbf{H}^L\mathbf{G}^{LN}\!=\!(\nabla_{\theta}\mathbf{H}^L\!-\!\nabla_{\theta}\mathbf{M}^{LN})$. In this way, we can equivalently rewrite the NTK with layer normalization (\textit{i.e.}, Eqn.~\ref{eqn:ntk_ln}) as,
\begin{equation}
    \begin{split}
    \mathbf{K}^{LN}
     &=\mathbf{A}^{LN}(\nabla_{\theta}\mathbf{H}^L-\nabla_{\theta}\mathbf{M}^{LN})^{\top}(\nabla_{\theta}\mathbf{H}^L-\nabla_{\theta}\mathbf{M}^{LN})\\
     &= \mathbf{A} ^{LN}(\mathbf{G}^{LN})^{\top} (\nabla_{\theta}\mathbf{H}^L)^{\top}\nabla_{\theta}\mathbf{H}^L \mathbf{G}^{LN}\\
     &= \mathbf{A}^{LN}(\mathbf{G}^{LN})^{\top} \mathbf{K} \mathbf{G}^{LN}.\\
    \end{split}
    \label{eq-2-2}
\end{equation}
Thus we can numerically calculate each element in $\mathbf{K}^{LN}$ as (here we omit the remainder term of $ \mathcal{O}(\sqrt{N})$):
\begin{equation}
    \begin{split}
\mathbf{K}^{LN}_{i,j,s,t}\!\approx\! \begin{cases} 
    \frac{N\kappa_1}{\sigma_t\sigma_s}(\frac{C\!-\!1}{C}\!-\!\frac{\sum^C_m \textbf{H}_{m,s}\textbf{H}_{m,t}}{C\sigma_t\sigma_s}),     &  for\ \  i=j,     s=t    \\
    \frac{N\kappa_1}{\sigma_t\sigma_s}(-\frac{1}{C}\!-\!\frac{\sum^C_m \textbf{H}_{m,s}\textbf{H}_{m,t}}{C\sigma_t\sigma_s}),     &  for\ \  i=j,     s\neq t\\
    \frac{N\kappa_2}{\sigma_t\sigma_s}(\frac{C\!-\!1}{C}\!-\!\frac{\sum^C_m \textbf{H}_{m,s}\textbf{H}_{m,t}}{C\sigma_t\sigma_s}),                         &  for\ \  i\neq j, s=t    \\
    \frac{N\kappa_2}{\sigma_t\sigma_s}(-\frac{1}{C}\!-\!\frac{\sum^C_m \textbf{H}_{m,s}\textbf{H}_{m,t}}{C\sigma_t\sigma_s}),                          &  for\ \  i\neq j, s\neq t
    \end{cases}.
    \end{split}
    \label{eqn:k_ln}
\end{equation}
According to the obtained explicit element values of $\mathbf{K}^{LN}$, the statistical values of NTK eigenvalues, namely, mean value $m_\lambda$, second momentum $s_\lambda$, variance $v_\lambda$, and the maximum eigenvalue $\lambda_{max}$, could be calculated directly,
\begin{equation}
\begin{aligned}
&\text{LN-based MLP:}\\
&\qquad m_\lambda \sim \frac{NC\sum_{i=1}^T\!\frac{\kappa_1}{\sigma_i^2}\!+\!\mathcal{O}(\sqrt{N})CT}{CT} \sim \mathcal{O}(N\!+\!\sqrt{N})\\
&\qquad s_\lambda \sim \mathcal{O}\left(C(N\!+\!\sqrt{N})^2\!+\!(CT\!-\!C)N\right) \\
&\qquad v_\lambda \sim \mathcal{O}\left((N\!+\!2\sqrt{N}\!+\!T)CN\right) \\
&\qquad \mathcal{O}(NC\!+\!\sqrt{N}C\!+\!CT) \leq \lambda_{max} \leq  \mathcal{O}(NC\sqrt{T}\!+\!\sqrt{N}CT).
\end{aligned}
\label{eqn:ntk_cmp_ln}
\end{equation}

\noindent \textbf{Layer normalization alleviates the spectral bias.} 
Comparing these statistical values without and with layer normalization, it is observed that the mean values are still slighted changed and remain the same scale of $\mathcal{O}(N\!+\!\sqrt{N})$. 
Nevertheless, the variances are manipulated from $\mathcal{O}\left((N\!+\!2\sqrt{N}\!+\!C)TN\right)$ to $\mathcal{O}\left((N\!+\!2\sqrt{N}\!+\!C)TN\right)$, considering the fact that the number of samples $T$ is largely greater than the number of attributes $C$, thus the variance is significantly reduced with the coefficient of $\frac{T}{C}$.
Moreover, the upper bound of the largest eigenvalue is also significantly reduced with an approximate factor of $\mathcal{O}(\sqrt{C})$. Once again, these alters of statistical values indicates a shift of NTK eigenvalues' distribution from a lower one to a higher one (as validated in Fig.~\ref{pic1}\textcolor{red}{a}). 
As a consequence, layer normalization can also alleviate the issue of the spectral bias to some extend through manipulating the pathological distribution of NTK eigenvalues.

\begin{figure*}[t]
  \centering
  \includegraphics[width=0.98\linewidth]{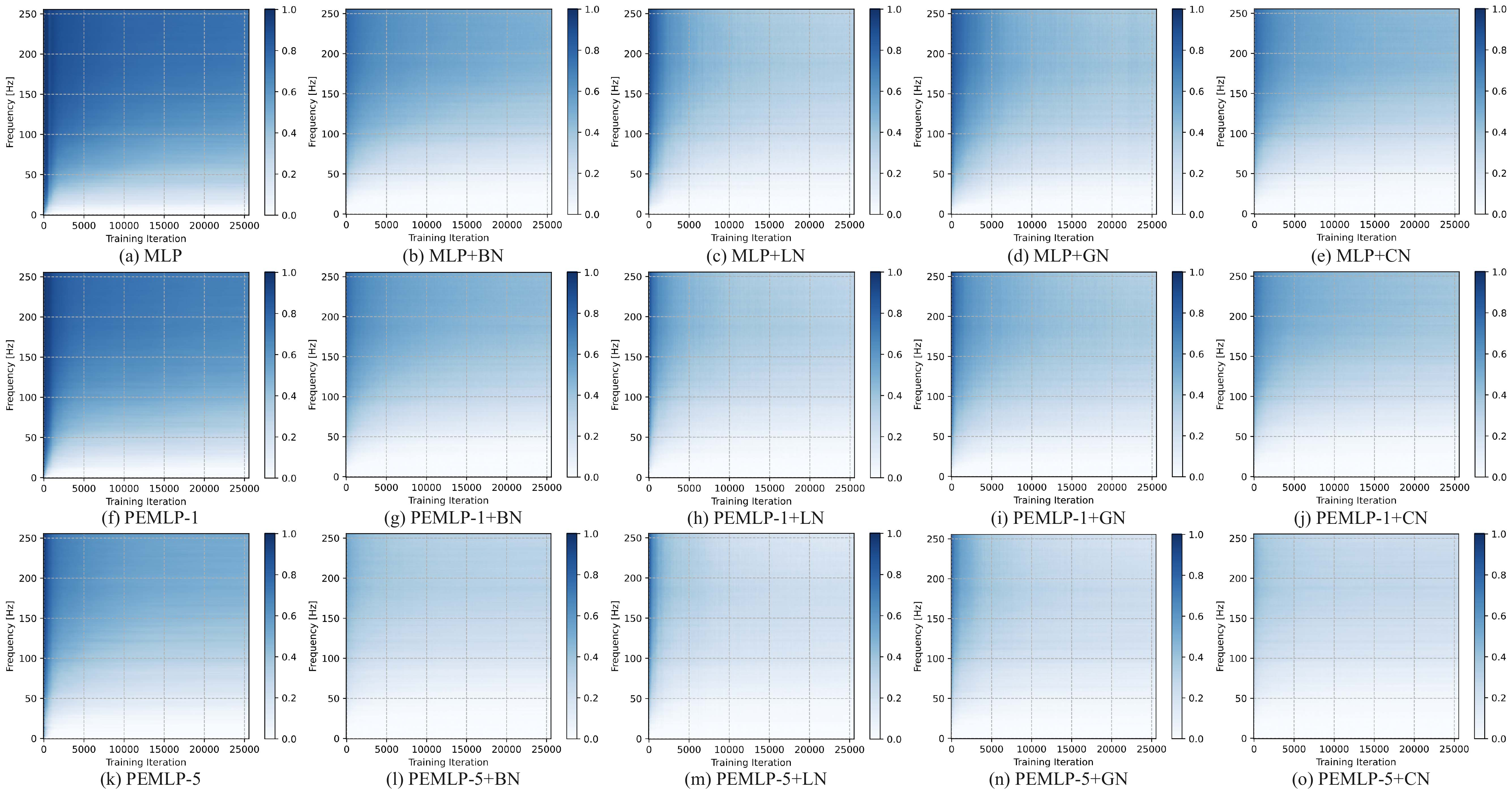}
  \vspace{-1em}
  \caption{Evolution of frequency-specific approximation error with training iterations of five different methods (x-axis for training iteration, y-axis for frequency and colormap for relative approximation error). Deeper color represents larger frequency-error.}
  \label{freq-error}
  \vspace{-1em}
\end{figure*}

\subsection{Explore the Joint Normalizations along Batch and Channel Dimensions}
As can be intuitively analyzed from Eqn.~\ref{eq-5} that, the effectiveness of BN is straightforwardly attributed to the mean subtraction-related projector $\mathbf{G}^{BN}$ and variance division-related projector $\mathbf{A}^{BN}$, which makes each sample attend to other samples along the batch dimension, thus correspondingly adjusting the mathematical representation of the NTK and manipulating the distribution of NTK eigenvalues.
Similarly, the function of LN also originates from the mean subtraction-related projector $\mathbf{G}^{LN}$ and variance division-related projector $\mathbf{A}^{LN}$ as deduced in Eqn.~\ref{eq-2-2}, in which case these projectors make samples in certain channel respond to other channels along the channel dimension and modify the mathematical expression of NTK accordingly.
Based on the observation of the theoretical analysis, we raise two new normalization methods with different combination approaches and \emph{normalization scopes}, namely, Global Normalization (GN) and Cross Normalization (CN).
These new normalization methods allow certain pixels to attend to pixels in other locations by applying normalization to both batch and channel dimensions jointly, which is expected to further optimize the mathematical form of the NTK and deliver better representational results for coordinate networks.

\noindent\textbf{Global Normalization}:
First we introduce the GN, for which each pixel attends to all the other pixels within the global normalization scope, as shown in Fig.~\ref{pic-norms}\textcolor{red}{c}.
The expression of GN can be written as,
\vspace{-0.5em}
\begin{equation}
\begin{aligned}
    &f^{GN}(\mathbf{X};\theta)_{i,t}=\frac{{\mathbf{H}_{i,t}}^{L}-{\mu}}{\sigma}\gamma_i + \beta_i\\
    &\mu = \frac{1}{CT}\sum^C_{m=1}\sum^T_{n=1}\mathbf{H}^{L}_{m,n},\\
    &\sigma=\sqrt{\frac{1}{CT}\sum^C_{m=1}\sum^T_{n=1}(\mathbf{H}^{L}_{m,n})^2-(\frac{1}{CT}\sum^C_{m=1}\sum^T_{n=1}\mathbf{H}^{L}_{m,n})^2}.
\end{aligned}
\label{eqn:gn}
\end{equation}

\noindent\textbf{Cross Normalization}:
We also propose CN to conduct the joint normalization along both batch and channel dimensions in another approach.
As shown in Fig.~\ref{pic-norms}\textcolor{red}{d}, CN allows certain pixel to attend to all other pixels within the corresponding channel and batch dimensions, delivering a cross-shaped normalization scope.
The expression of CN can be written as,
\vspace{-0.5em}
\begin{equation}
\begin{aligned}
    &f^{CN}(\mathbf{X};\theta)_{i,t}=\frac{{\mathbf{H}_{i,t}}^{L}-{\mu_{i,t}}}{\sigma_{i,t}}\gamma_i + \beta_i\\
    &\mu_{i,t} = \frac{1}{C+T}(\sum^C_{m=1}\mathbf{H}^{L}_{m,t}+\sum^T_{n=1}\mathbf{H}^{L}_{i,n}),\\
    &\sigma_{i,t}=\sqrt{\frac{1}{C+T}(\sum^C_{m=1}(\mathbf{H}^{L}_{m,t})^2+\sum^T_{n=1}(\mathbf{H}^{L}_{i,n})^2)-\mu_{i,t}^2}.
\end{aligned}
\label{eqn:cn}
\end{equation}

In summary, batch normalization has a \emph{normalization scope} of $T$ (the term normalization scope is utilized to describe the range within which normalization takes effect and certain pixel interacts with other pixels), and layer normalization has a normalization scope of $N$. We aim to enlarge the normalization scope and make each pixel attend to more other pixels, thus further improving the NTK eigenvalue distribution. Therefore, we introduce Global Normalization with a extremely large normalization scope of $T\times N$, and Cross Normalization with a larger scope of $T+N-1$, which is moderate but still comprehensive.
To better distinguish the joint property of GN and CN, the term \emph{normalization degree} is utilized to describe the number of independent sets of pixels which different pixels attend to, in other words, the amount of the mean values $\mu$ and variance values $\sigma$.
For the cases of BN and LN, the normalization degrees are $N$ and $T$, respectively.
For GN, it has a extremely low normalization degree of $1$, thus pixels in different locations share the same mean $\mu$ and variance $\sigma$. 
This low normalization degree maintains a low computational complexity while may hinder further performance enhancement due to the excessive coupling between different pixels. Nonetheless, it still performs significantly better than model without normalization.
For CN, it has a high normalization degree of $T\times N$, thus each pixel in different locations has unique mean and variance values. This allows for dynamic adjustments for individual pixels, achieving a better trade-off between normalization scope and degree, and promising the enhanced performance\footnote{Deriving the eigenvalues' distribution of CN and GN-based coordinate networks proves to be a difficult task due to the highly complexity of the mixed operations of BN and LN, as a result, we only provide the results of applying them to verify the efficacy throughout the paper.}.

\subsection{Discussion}
\subsubsection{Compatibility with Fourier Feature-based NTK manipulation}
Apart from the normalization techniques, positional encoding-based MLP (PEMLP)~\cite{tancik2020fourier} has been proved a successful method in alleviating the spectral-bias of coordinate network. According to NTK analysis in \cite{tancik2020fourier}, since the Fourier-based positional encoding only maps the low-dimensional input coordinates to high-dimensional Fourier feature space without affecting the internal architecture of networks, thus the role of positional encoding could be viewed as adding an additional term $h_{\gamma}$ to the original NTK, \textit{i.e.}, a composed NTK function $\mathbf{K}\circ h_{\gamma}$, where $h_{\gamma}$ is a linear combination of the used Fourier bases and is independent of the subsequent network. As a result, the mean and variance of the PEMLP's NTK will also change following the rule summarized in Eqns.~\ref{eqn:ntk_cmp_bn} and \ref{eqn:ntk_cmp_ln}, thus the `frequency-specified spectral bias' in PEMLP could also be alleviated by normalization techniques. 

\vspace{0.5em}
\subsubsection{Simple examples for fitting a 1D signal and a 2D image. } 
Fig.~\ref{pic1}\textcolor{red}{a} compares the NTK's eigenvalues without and with normalization techniques for learning a 1D signal. 
In the standard MLP, it is noticed that the peak probability appears in $\sim 10^{-3}$. When the normalization is applied, the distribution is significantly changed and appears as a whole shift from lower one to higher one. For example, the peak probability is shifted to $\sim 10^{0}$, $10^{-1}$, $\sim 10^{-2}$ and $\sim 10^{0.1}$ when the BN, LN, GN and CN are applied, respectively, meanwhile the largest eigenvalues are also reduced from $\sim10^{6}$ to $\sim 10^{5}$, validating the theoretical results analysed above. 
Furthermore, we also count the eigenvalues of NTK using classical positional encoding~\cite{tancik2020fourier} with 1 and 5 Fourier bases. As shown in Fig.~\ref{pic1}\textcolor{red}{a}, normalization techniques could provide similar ability with PEMLP for manipulating the eigenvalues. 

Fig.~\ref{pic1}\textcolor{red}{b} compares the results for fitting a 2D image. It is noticed that the performance of the reconstructed image also follows similar rule, \textit{i.e.}, more high-frequency details are reconstructed when normalization techniques are used. Fig.~\ref{freq-error} visualizes the frequency-specific training dynamics of MLP/PEMLP with different normalization techniques. In ReLU-based MLP, the spectral bias appears obviously and the training error of high-frequency components reduced slowly. PE technique alleviates the spectral-bias but the training error of high-frequency components still reduces slowly and remains at $\sim 0.4$. By applying the normalization techniques, especially the cross normalization, the training of high-frequency components converges at a very early stage and the training error reduces to $<0.2$, verifying the theoretical analysis and the efficacy of the proposed cross normalization. In summary, 
\begin{prop}
    Normalization techniques alleviate the spectral bias by making a shift of NTK's eigenvalues distribution from a lower one to a higher one.
\end{prop}

\begin{table*}[!t]
\centering
\setlength\tabcolsep{3pt}
\caption{Results of different coordinate networks on the task of 2D image compression. The results are measured in PSNR and SSIM. We color code each cell as \colorbox{c1}{best}, \colorbox{c2}{second best}, and \colorbox{c3}{third best}.}
\begin{tabular}{ll|ccccccccccccc} 

\toprule
Metrics               & Model         & SIREN & MFN   &WIRE   & ReLU  & ReLU+BN & ReLU+LN & ReLU+GN & ReLU+CN & PE    & PE+BN & PE+LN & PE+GN & PE+CN \\
\midrule
\multirow{5}{*}{PSNR} &5$\times$20    & 23.18 & 22.03 & 21.95 & 21.21 &  23.39  &  23.36  &  23.02  & 23.55   & 22.10 & \cellcolor{c3}24.26 & \cellcolor{c2}24.42 & 24.25 & \cellcolor{c1}24.51 \\  
                      &5$\times$30    & 24.57 & 23.34 & 23.10 & 22.57 &  24.39  &  24.22  &  24.09  & 24.62   & 23.47 & 25.54 & \cellcolor{c2}25.84 & \cellcolor{c3}25.56 & \cellcolor{c1}25.92 \\    
                      &10$\times$28   & 25.52 & 25.25 & 25.05 & 23.25 &  26.27  &  25.96  &  25.58  & 26.49   & 24.63 & \cellcolor{c2}26.85 & \cellcolor{c3}26.61 & 26.36 & \cellcolor{c1}26.91 \\    
                      &10$\times$40   & 28.32 & 26.92 & 27.34 & 23.56 &  27.95  &  27.57  &  27.02  & 28.21   & 26.66 & \cellcolor{c2}28.71 & \cellcolor{c3}28.45 & 28.06 & \cellcolor{c1}28.97 \\    
                      &13$\times$49   & \cellcolor{c3}29.42 & 28.14 & 29.02 & 24.33 &  29.08  &  28.26  &  27.86  & 29.22   & 28.23 & \cellcolor{c2}30.84 & 29.38 & 29.37 & \cellcolor{c1}31.08 \\ 
\midrule
\multirow{5}{*}{SSIM} &5$\times$20    & 0.540 & 0.482 & 0.466 & 0.506 &  0.558  &  0.552  &  0.546  & 0.561   & 0.535 & 0.577 & \cellcolor{c2}0.584 & \cellcolor{c3}0.581 &\cellcolor{c1} 0.586 \\  
                      &5$\times$30    & 0.594 & 0.544 & 0.513 & 0.510 &  0.589  &  0.587  &  0.568  & 0.593   & 0.562 & 0.629 & \cellcolor{c2}0.637 & \cellcolor{c3}0.633 & \cellcolor{c1}0.640 \\    
                      &10$\times$28   & 0.637 & 0.587 & 0.565 & 0.521 &  0.650  &  0.637  &  0.627  & 0.654   & 0.590 & \cellcolor{c2}0.682 & \cellcolor{c3}0.669 & 0.667 & \cellcolor{c1}0.685 \\    
                      &10$\times$40   & 0.712 & 0.626 & 0.681 & 0.533 &  0.693  &  0.686  &  0.670  & 0.700   & 0.626 & \cellcolor{c2}0.754 & \cellcolor{c3}0.737 & 0.734 & \cellcolor{c1}0.760 \\    
                      &13$\times$49   & 0.727 & 0.715 & 0.743 & 0.540 &  0.754  &  0.727  &  0.713  & 0.757   & 0.649 & \cellcolor{c2}0.821 & \cellcolor{c3}0.779 & \cellcolor{c3}0.779 &\cellcolor{c1}0.827 \\ 
\bottomrule
\end{tabular}
\label{image-results}
\end{table*}

\begin{figure*}[!t]
  \centering
  \includegraphics[width=\linewidth]{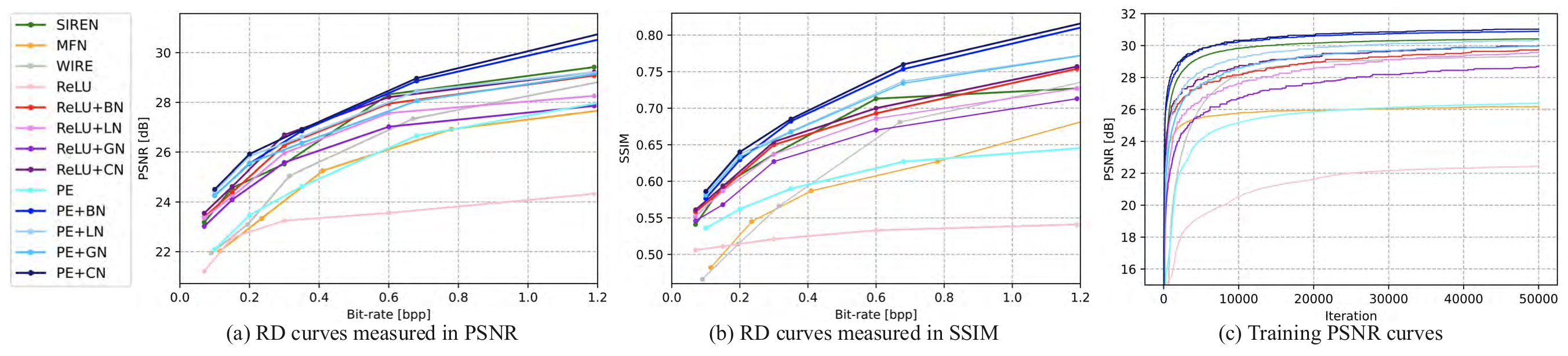}
  \caption{(a) Rate-distortion curves measured in PSNR of various coordinate networks under different bpps trained on the Kodak dataset. (b) Rate-distortion curves measured in SSIM of various coordinate networks under different bpps trained on the Kodak dataset. (c) exhibits the training curves of these methods on 2D image.}
  \label{image-curves}
\end{figure*}

\begin{figure*}[!t]
  \centering
  \includegraphics[width=\linewidth]{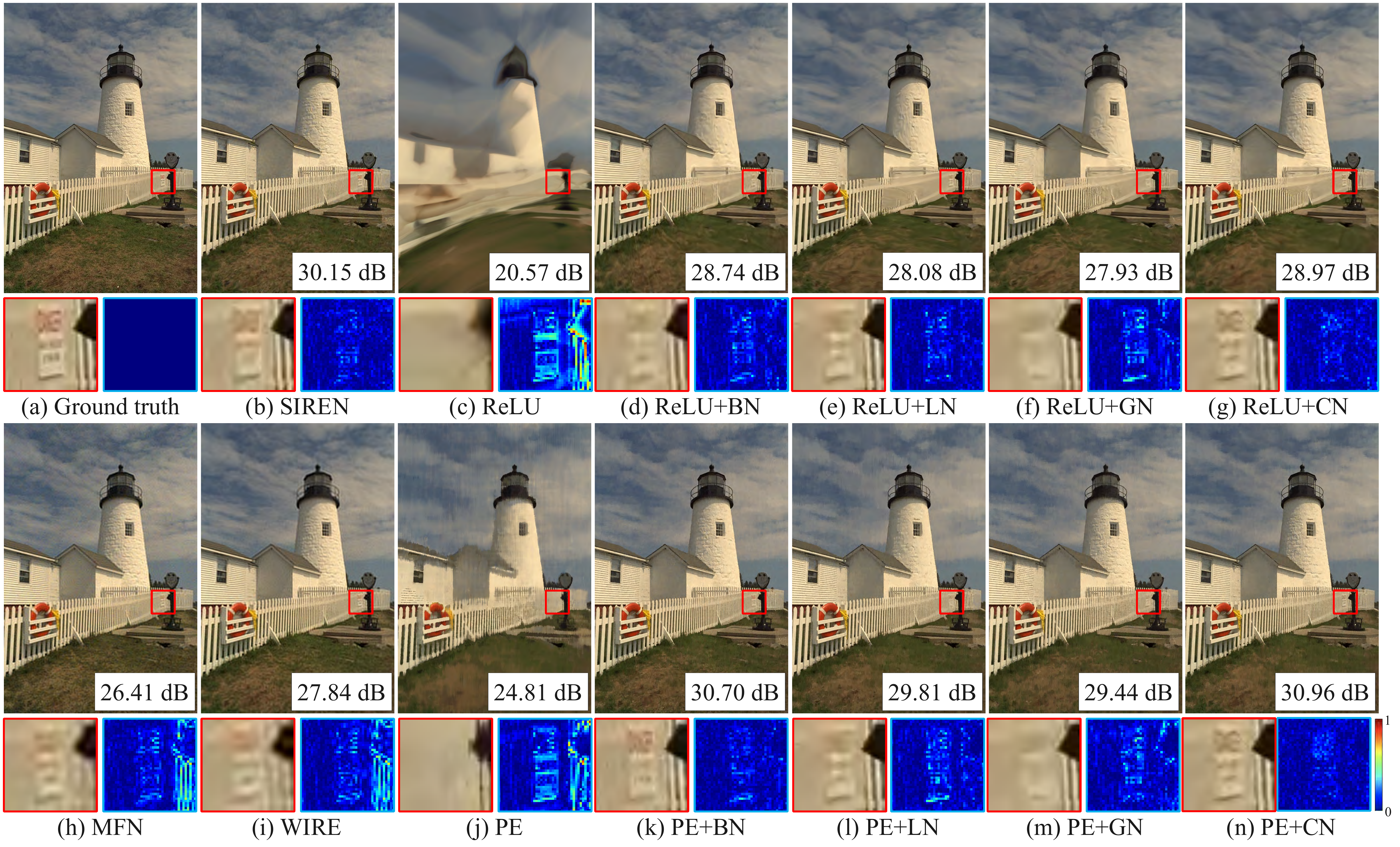}
  \caption{Comparisons of different methods for representing the 2D Image Tower. The corresponding Fourier spectra are also visualized.}
  \label{lena}
\end{figure*}

\section{Experiments}
In this section, we first validate the effectiveness of applying the four normalization techniques (namely, BN, LN, GN and CN) to coordinate networks on four representation and simple optimization, \textit{i.e.}, 2D image representation and compression, 3D shape representation, 2D computed tomography (CT) reconstruction, and 3D magnetic resonance imaging (MRI) reconstruction.

\textbf{Compared methods.} 
For these four tasks, a total of thirteen methods are compared, including the network with traditional activation ReLU (ReLU)~\cite{nair2010rectified}, ReLU based network with Fourier features positional encoding (PE)~\cite{tancik2020fourier},  network with sinusoidal nonlinearity (SIREN)~\cite{sitzmann2020implicit}, multiplicative filter network (MFN)~\cite{fathony2020multiplicative} and complex Gabor wavelet nonlinearity (WIRE)~\cite{saragadam2023wire}. Because these four normalization techniques are a universal tool, the results of applying these normalizations on the ReLU (labeled as ReLU+BN, ReLU+LN, ReLU+GN, and ReLU+CN) and positional encoding ReLU (labeled as  PE+BN, PE+LN, PE+GN, and PE+CN) are all compared. 
For these methods equipped with normalization, we add one corresponding normalization layer before the ReLU activation of each fully-connected layer. 
As a common practice, the encoding scale of PE is set as $10$~\cite{mildenhall2021nerf}, the frequency parameter $\omega_0$  of SIREN is set as $30$~\cite{sitzmann2020implicit}, the frequency parameter $\omega$ and the spread parameter $s$ of WIRE are respectively set as $20$ and $10$~\cite{saragadam2023wire}.
The weights of all the networks are randomly initialized. For SIREN and MFN, we utilize the specific weight initialization schemes as raised in \cite{sitzmann2020implicit} and \cite{fathony2020multiplicative}, respectively. For the left eleven methods, we utilize the default LeCun random initialization~\cite{lecun2002efficient}.

\begin{table*}[!t]
\centering
\setlength\tabcolsep{3pt}
\caption{Results of different coordinate networks on 3D Shape Representation task. The results are measured in IoU. We color code each cell as \colorbox{c1}{best}, \colorbox{c2}{second best}, and \colorbox{c3}{third best}.}
\begin{tabular}{l|ccccccccccccc} 
\toprule
Scene         & SIREN & MFN   & WIRE  & ReLU  & ReLU+BN & ReLU+LN & ReLU+GN & ReLU+CN & PE    & PE+BN & PE+LN & PE+GN & PE+CN \\
\midrule
Armadillo     & 0.950 & 0.983 & 0.969 & 0.976 & 0.981   & 0.982   & 0.983   & 0.986   & \cellcolor{c3}0.989 & \cellcolor{c2}0.996 & \cellcolor{c2}0.996 & \cellcolor{c1}0.997 & \cellcolor{c1}0.997 \\ 
Budda         & 0.976 & 0.969 & 0.980 & 0.968 & 0.979   & 0.983   & 0.978   & 0.987   & \cellcolor{c3}0.988 & \cellcolor{c1}0.997 & \cellcolor{c2}0.996 & \cellcolor{c1}0.997 & \cellcolor{c1}0.997 \\ 
Bunny         & 0.947 & 0.992 & 0.989 & 0.991 & 0.994   & 0.994   & 0.994   & 0.995   & \cellcolor{c3}0.997 & \cellcolor{c1}0.999 & \cellcolor{c2}0.998 & \cellcolor{c1}0.999 & \cellcolor{c1}0.999 \\
Dragon        & 0.970 & 0.974 & 0.975 & 0.953 & 0.981   & 0.982   & 0.981   & \cellcolor{c3}0.987   & 0.986 & \cellcolor{c2}0.996 & \cellcolor{c2}0.996 & \cellcolor{c2}0.996 & \cellcolor{c1}0.998 \\
Asian Dragon  & 0.937 & 0.969 & 0.942 & 0.938 & 0.961   & 0.960   & 0.964   & 0.970   & 0.978 & 0.985 & \cellcolor{c3}0.990 & \cellcolor{c2}0.994 & \cellcolor{c1}0.995 \\
Drill         & 0.993 & \cellcolor{c2}0.998 & \cellcolor{c3}0.997 & 0.966 & 0.993   & 0.994   & 0.994   & 0.995   & \cellcolor{c1}0.999 & \cellcolor{c1}0.999 & \cellcolor{c3}0.997 & \cellcolor{c3}0.997 & \cellcolor{c1}0.999 \\
Lucy          & 0.957 & 0.971 & 0.963 & 0.924 & 0.964   & 0.973   & 0.966   & 0.977   & 0.985 & \cellcolor{c2}0.994 & \cellcolor{c3}0.993 & \cellcolor{c2}0.994 & \cellcolor{c1}0.996 \\
Thai Statue   & 0.964 & 0.972 & 0.964 & 0.904 & 0.958   & 0.966   & 0.965   & 0.971   & 0.968 & \cellcolor{c3}0.992 & 0.\cellcolor{c2}994 & \cellcolor{c3}0.992 & \cellcolor{c1}0.995 \\
\midrule
Average       & 0.962 & 0.978 & 0.972 & 0.952 & 0.976   & 0.979   & 0.978   & 0.985   & 0.986 & \cellcolor{c3}0.995 & \cellcolor{c3}0.995 & \cellcolor{c2}0.996 & \cellcolor{c1}0.997 \\
\bottomrule
\end{tabular}
\label{shape-results}
\end{table*}

\begin{figure*}[!t]
  \centering
  \includegraphics[width=\linewidth]{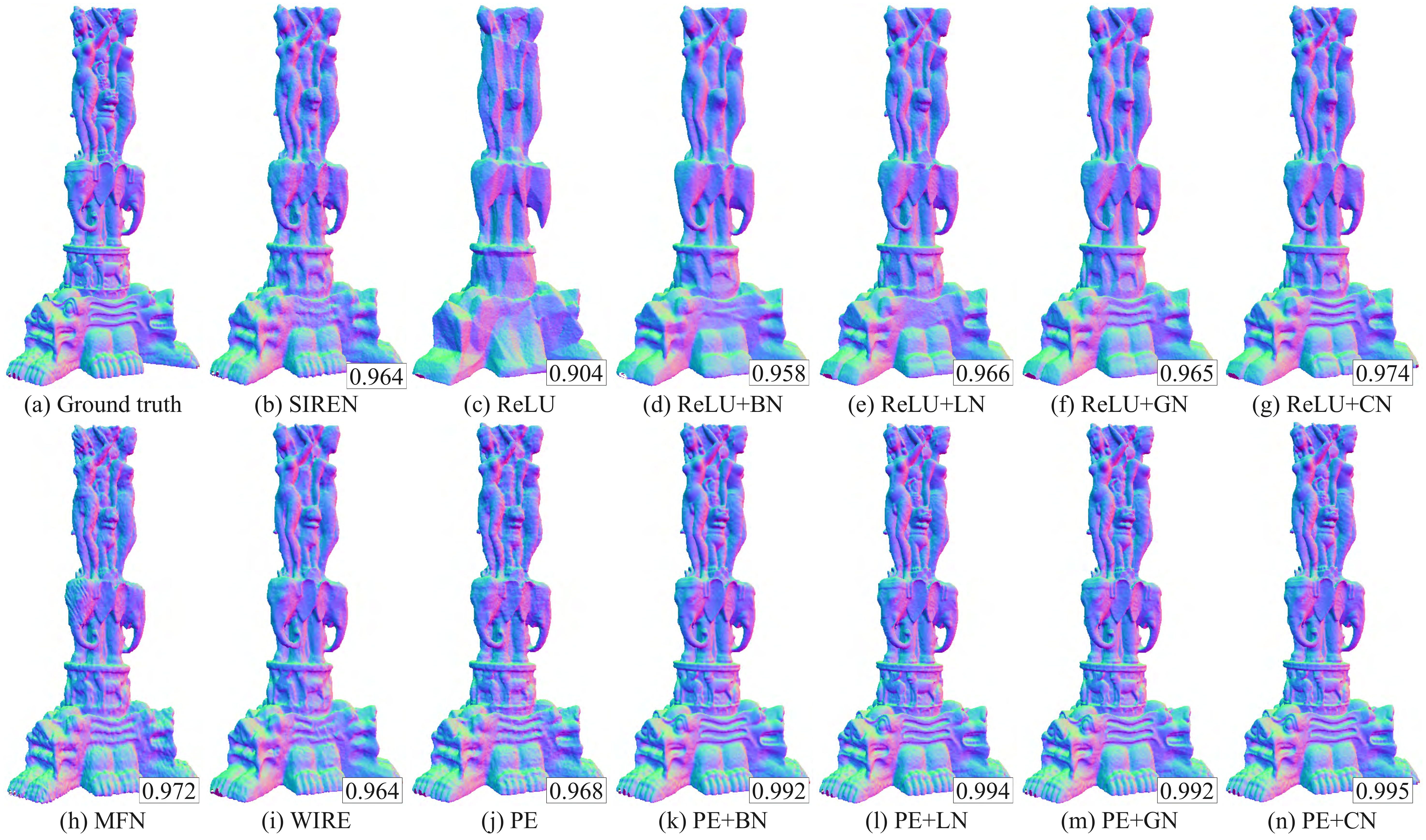}
  \caption{Meshes generated with occupancy volumes by various coordinate networks.}
  \label{sdf-vis}
\end{figure*}

\subsection{2D Image Compression}
\textbf{Configurations.}
We first use an image compression task to evaluate the performance of applying normalization techniques to the coordinate networks. 
In this task, the coordinate network aims at learning a 2D function $f: \mathbb{R}^2\mapsto\mathbb{R}^3$ with 2D pixel location input and 3D RGB color output. 
We perform experiments on the Kodak image dataset ~\cite{dupont2021coin} consisting of 24 RGB images with a high resolution of $768\times512$.
To utterly explore the representational capacity of various methods, we use networks with limited parameters as introduced in ~\cite{dupont2021coin}, namely, networks with the configurations of (in the format of [$hidden\  layers \times hidden\  features$]) [5$\times$20], [5$\times$30], [10$\times$28], [10$\times$40], and [13$\times$49]. 
Thus it can be regarded as the image compression task using coordinate networks.
We use the $L_2$ distance between the network output and the ground truth as the loss function. 
All the models are trained for 100,000 iterations using Adam optimizer~\cite{kingma2014adam}. The initial learning rate is set as $2e\!-\!4$, except for these method with normalization, which use a larger initial learning rate of $1e\!-\!2$ thanks to the insensitivity to the learning rate of normalization techniques. The batch size is equal to the number of image pixels.

\textbf{Results.}
Tab.~\ref{image-results} exhibits the average PSNR (Peak Signal-to-Noise Ratio) and SSIM (Structural Similarity Index) of these methods. 
Accordingly, we plot the rate-distortion (RD) curves of these methods under various bpps (bits-per-pixel = number-of-parameters$\times$bits-per-parameter / number-of-pixels) measured by PSNR and SSIM as shown in Fig.~\ref{image-curves}\textcolor{red}{a} and Fig.~\ref{image-curves}\textcolor{red}{b}, respectively.
As can be observed, ReLU consistently presents the worst performance due to the spectral bias, while applying normalizations can significantly enhance the performance of networks with ReLU. For example, BN, LN, GN, and CN improve the PSNR of ReLU by up to $4.75$dB, $3.93$dB, $3.53$dB, and $4.89$dB with the network of [13$\times$49], respectively. 
All the normalizations also exhibit competitive performance compared to other existing method for that they consistently obtain higher PSNR than PE (up to $1.64$dB, $1.33$dB, $0.95$dB, and $1.86$dB with the network of [10$\times$28] for BN, LN, GN, and CN, respectively), WIRE (up to $1.44$dB, $1.41$dB, $1.08$dB, $1.60$dB, with the network of [5$\times$20] for BN, LN, GN, and CN, respectively) and MFN (up to $1.36$dB, $1.33$dB, $1.00$dB, $1.52$dB, with the network of [5$\times$20] for BN, LN, GN, and CN, respectively).
Besides, BN, LN , and CN also surpass SIREN with the network of [5$\times$20] and [10$\times$28].
Compared these normalization techniques, CN consistently achieves the best performance, demonstrating the superiority of normalization in a cross way.
Moreover, normalizations can also notably promote the performance of PE. 
For example, BN, LN, GN, and CN improve the PSNR of PE by up to $2.61$dB, $1.15$dB, $1.14$dB, and $2.85$dB with the network of [13$\times$49], respectively.
It is worth emphasizing that PE+CN consistently achieves the highest PSNR among all the thirteen methods, which further validates the effectiveness of CN.

\textbf{Training dynamics.}
To explore the training dynamics of networks with normalizations and other methods, we plot the training PSNR curves of these methods fitted on 2D image as shown in Fig.~\ref{image-curves}\textcolor{red}{c}. Normalizations not only accelerate the convergence speed of the ReLU (as well as PE) based coordinate networks, but also significantly enhance the PSNR. In comparison with other methods, normalizations continue to exhibit an advantage in terms of convergence speed.

\begin{table*}[!t]
\centering
\setlength\tabcolsep{3pt}
\caption{Results of different coordinate networks on 2D CT reconstruction task. The results are measured in PSNR and SSIM. We color code each cell as \colorbox{c1}{best}, \colorbox{c2}{second best}, and \colorbox{c3}{third best}.}
\begin{tabular}{ll|ccccccccccccc} 
\toprule
Metrics               & Model         & SIREN & MFN   &WIRE   & ReLU  & ReLU+BN & ReLU+LN & ReLU+GN & ReLU+CN & PE    & PE+BN & PE+LN & PE+GN & PE+CN  \\
\midrule
\multirow{4}{*}{PSNR} &2$\times$128   & 28.30 & 24.82 & 28.30 & 26.31 & 28.50   & 27.55   & 27.07   & 28.72   & 28.16 & \cellcolor{c2}31.14 & \cellcolor{c3}30.08 & 29.71 & \cellcolor{c1}31.44  \\  
                      &2$\times$256   & 16.16 & 27.97 & 28.26 & 26.78 & 28.41   & 27.75   & 27.40   & 28.77   & 28.11 & \cellcolor{c2}31.60 & \cellcolor{c3}30.52 & 30.03 & \cellcolor{c1}31.71  \\    
                      &4$\times$128   & 18.31 & 29.15 & 28.31 & 28.21 & 30.42   & 29.46   & 28.75   & 30.81   & 30.42 & \cellcolor{c2}32.45 & \cellcolor{c3}32.34 & 31.61 & \cellcolor{c1}32.62  \\    
                      &4$\times$256   & 18.32 & 30.92 & 28.70 & 28.30 & 30.48   & 29.67   & 28.67   & 30.87   & 30.00 & \cellcolor{c2}32.64 & \cellcolor{c3}32.47 & 31.86 & \cellcolor{c1}32.87  \\    
\midrule
\multirow{4}{*}{SSIM} &2$\times$128   & 0.803 & 0.527 & 0.756 & 0.821 & 0.882   & 0.865   & 0.833   & 0.885   & 0.873 & \cellcolor{c2}0.913 & \cellcolor{c3}0.903 & 0.879 & \cellcolor{c1}0.915  \\  
                      &2$\times$256   & 0.198 & 0.677 & 0.729 & 0.841 & 0.884   & 0.856   & 0.848   & 0.886   & 0.890 & \cellcolor{c2}0.910 & \cellcolor{c3}0.908 & 0.890 & \cellcolor{c1}0.912  \\    
                      &4$\times$128   & 0.201 & 0.765 & 0.744 & 0.862 & 0.923   & 0.903   & 0.887   & 0.928   & 0.921 & \cellcolor{c2}0.934 & \cellcolor{c2}0.934 & \cellcolor{c3}0.930 & \cellcolor{c1}0.936  \\    
                      &4$\times$256   & 0.261 & 0.856 & 0.719 & 0.873 & 0.924   & 0.905   & 0.890   & \cellcolor{c3}0.930   & 0.917 & \cellcolor{c2}0.937 & \cellcolor{c2}0.937 & \cellcolor{c3}0.930 & \cellcolor{c1}0.939  \\    
\bottomrule
\end{tabular}
\label{ct-results}
\end{table*}

\begin{figure*}[!t]
  \centering
  \includegraphics[width=\linewidth]{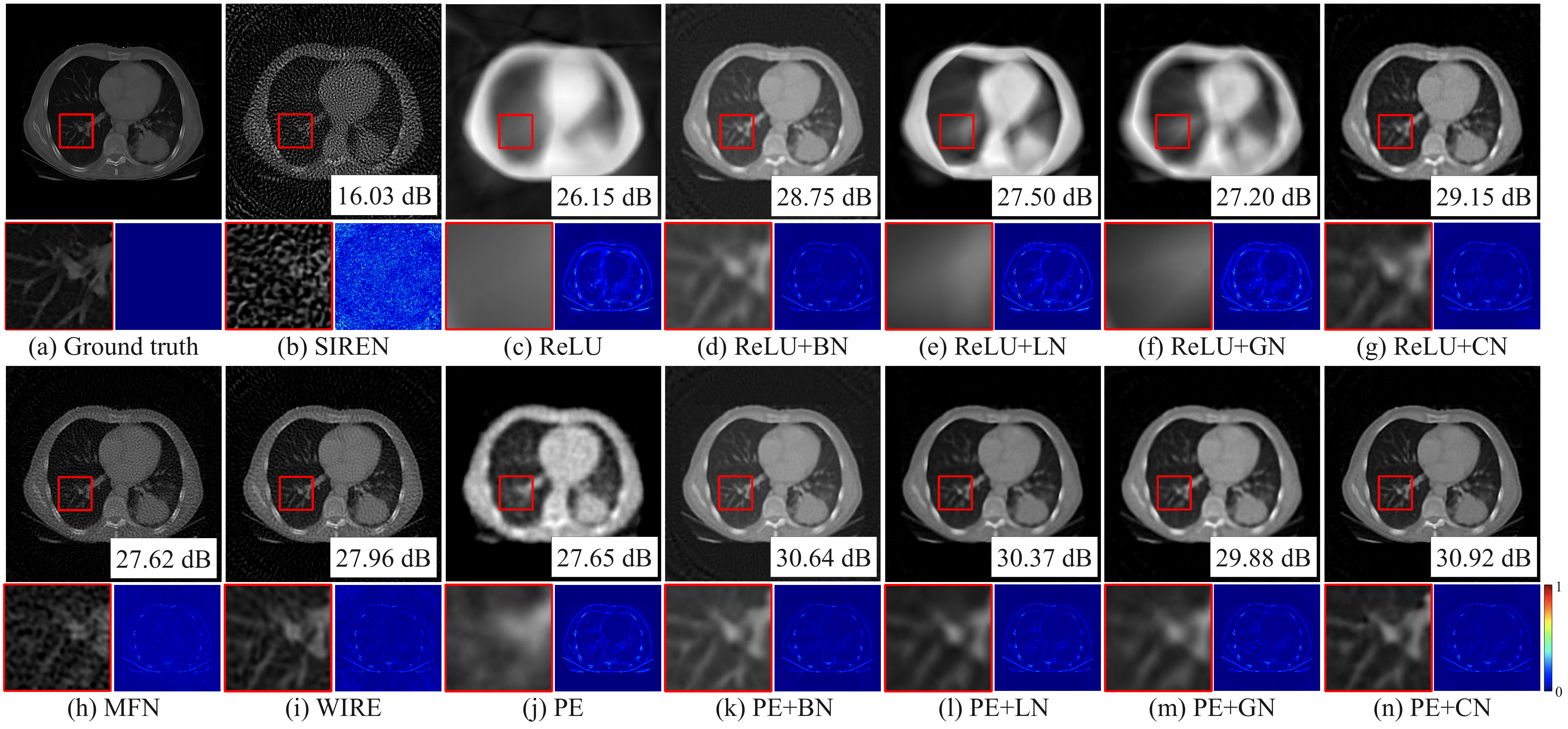}
  \caption{Comparisons of different coordinate networks for CT reconstruction. The corresponding error maps are also visualized.}
  \label{ct}
\end{figure*}

\textbf{Visualization.}
We visualize the reconstructed images and corresponding Fourier frequency spectra of various methods in Fig.~\ref{lena}. The target visual signal is 2D image Tower in the Kodak dataset, and the network is of [13$\times$49].
As can be obviously observed in Fig.~\ref{lena}\textcolor{red}{c}, the reconstructed image of ReLU is over-smoothed and poor-quality, failing to present the details of the target image. 
For PE, the quality of the reconstructed image (Fig.~\ref{lena}\textcolor{red}{j}) is poor for that the overall image is blurry and the finer details are failed to be restored. Moreover, both low- and high-frequencies are learned limitedly.
However, after applying normalizations to the networks, the quality of the reconstructed images (Fig.~\ref{lena}\textcolor{red}{d,e,f,g}, and Fig.~\ref{lena}\textcolor{red}{k,l,m,n}) is significantly improved, and the details are clearly represented. 
Furthermore, as clearly shown by the Fourier spectra, more low- and high-frequency components are learned effectively with the application of normalization, illustrating the effectiveness of our method on alleviating the spectral bias.

For SIREN (Fig.~\ref{lena}\textcolor{red}{b}), MFN (Fig.~\ref{lena}\textcolor{red}{h}) and WIRE (Fig.~\ref{lena}\textcolor{red}{i}), the presence of artifacts is clearly observable, resulting from over-fitting to high-frequency components. This, in turn, leads to noticeable noise and compromises the quality of reconstruction. 
This issue is particularly severe for MFN and WIRE, as indicated by their corresponding Fourier spectra, both methods tend to emphasize high frequencies while neglecting low-frequency components.
In contrast, the coordinate networks with normalizations produce clean background reconstructions with minimal artifacts and noise. This phenomenon further demonstrates the robustness of normalizations to noise and its superior capability in representing the target visual signal.

\subsection{3D Shape Representation}
\textbf{Configurations.}
In this section, we demonstrate the representational capacity of batch normalization for representing 3D shapes as occupancy networks. 
To be specific, the input data is a mesh grid with $512^3$ resolution, where the voxels inside the volume are assigned as $1$, and the voxels outside the volume are assigned as $0$.
Then we use the occupancy network to implicitly represent a 3D shape as the “decision boundary” of coordinate networks, which is trained to output 0 for points outside the shape and 1 for points inside the shape. 
In this way, the coordinate network learns a 3D mapping function $f: \mathbb{R}^3\mapsto\mathbb{R}^1$ with 3D voxel location input and 1D occupancy value output.
Test error is calculated using cross-entropy loss between the network output and the ground truth points.
We conduct the experiments on the data from Stanford 3D Scanning Repository~\cite{standord-3D-scanning}.
We use networks with the architecture configurations, of [2$\times$256]. 
In the experiment, all the models are trained for 200 epochs using Adam optimizer with a initial learning rate of $5e\!-\!3$ and a minimum learning rate of $5e\!-\!4$. 
100,000 points are randomly sampled in each iteration during the training process. The network outputs are extracted as a $512^3$ grid using marching cubes~\cite{lorensen1998marching} with a threshold of 0.5 for evaluation and visualization.

\textbf{Results.}
Tab.~\ref{shape-results} exhibits the experimental results evaluated by IoU (Intersection over Union).
As can be observed, normalizations significantly improve the performance of simple ReLU, averagely enhancing the IoU up to $2.52\%$, $2.84\%$, $2.73\%$ and $3.47\%$ by BN, LN, GN, and CN. respectively. 
In addition, normalizations also substantially boost the representation quality of PE, increasing the IoU by an average of up to $0.91\%$, $0.91\%$, $1.10\%$, and $1.12\%$ when added with BN, LN, GN, and CN, respectively.
Moreover, PE+CN achieves the best performance among the thirteen methods.

\textbf{Visualization.}
Fig.~\ref{sdf-vis} visualizes the meshes of Thai Statue scene represented by these methods.
As shown in Fig.~\ref{sdf-vis}, ReLU makes the surface over-smooth, failing to represent high-frequency components, while normalizations significantly improve such phenomenon allowing finer details to emerge, alleviating the issue of spectral bias. 
Besides, MFN introduces too many undesired textures and fluctuations on the surface, indicating over-fitting to high-frequencies and noise.
Furthermore, PE+CN achieves the best representational result, providing clear details without incurring artifacts or noise.

\subsection{2D Computed Tomography}
\textbf{Configurations.}
In the CT reconstruction task, we observe integral projections of a density field instead of direct supervisions. In our experiments, we train a network $f: \mathbb{R}^2\mapsto\mathbb{R}^1$ that takes in 2D pixel coordinates and predicts the corresponding volume density at each location.
We conduct the experiments on the x-ray colorectal dataset~\cite{saragadam2023wire,clark2013cancer}, each image has a resolution of $512\times 512$ and is emulated with 100 CT measurements.
We use networks with three architecture configurations, namely, [2$\times$128], [2$\times$256], and [4$\times$128]. 
To solve the inverse problem, the network is indirectly supervised by the MSE loss between a sparse set of ground-truth integral projections and integral projections computed from the network's output.
All the models are trained for 20,000 iterations using Adam optimizer with a initial learning rate of $5e\!-\!3$ and a minimal learning rate of $5e\!-\!4$. The batch size is equal to the number of input coordinates.

\textbf{Results.}
Tab.~\ref{ct-results} provides a quantitative comparison of the performance of various methods, measured in PSNR and SSIM. As can be observed, PE+CN consistently attains the highest PSNR and SSIM with all the network configurations, emphasizing the effectiveness of CN. 
Furthermore, simple applying CN to the network surpasses all the other existing methods except for MFN with the network of [4$\times$256], which has a slightly higher PSNR ($0.05$dB) while suffers from a much lower SSIM of $0.856$, significantly lagging behind the $0.930$ of CN.
In addition, adding ReLU with BN also achieves competitive results compared with other existing methods, exceeding SIREN, MFN, WIRE, and PE up to $0.20$dB, $3.68$dB, $0.20$dB, and $0.34$dB respectively with the network of [2$\times$128]. 
For MFN and WIRE, the SSIM has consistently been low, they are surpassed by PE+CN up to $22\%$ and $26\%$ respectively with the network of [4$\times$128].
While SIREN can only achieve moderate performance with the network of [2$\times$128], for other network configurations, SIREN fails to reconstruct the target signal effectively, delivering poor PSNR and SSIM.

\textbf{Visualization.}
Fig.~\ref{ct} visualizes the CT images reconstructed by different methods with the network of [2$\times$256], the corresponding error maps are also visualized in the bottom right corner of each method.
As can be observed, ReLU (Fig.~\ref{ct}\textcolor{red}{c}) leads to excessively smooth results, only displaying blurry patterns. The result of PE (Fig.~\ref{ct}\textcolor{red}{j}) is noisy, failing to exhibit the precise details. When applying normalization to ReLU and PE, the reconstruction quality (Fig.~\ref{ct}\textcolor{red}{d,g} and Fig.~\ref{ct}\textcolor{red}{k,l,m,n}) is significantly improved as the finer details are effectively represented.
The other methods of novel nonlinear activations (SIREN, MFN, and WIRE) all result in obvious artifacts, especially for SIREN, which achieves a extremely low PSNR of only $16$dB. This phenomenon may be caused by the over-fitting of these methods to the high frequencies, thus introducing tremendous noise.

\begin{table*}[!t]
\centering
\setlength\tabcolsep{3pt}
\caption{Results of different coordinate networks on 3D MRI reconstruction task. The results are measured in PSNR and SSIM. We color code each cell as \colorbox{c1}{best}, \colorbox{c2}{second best}, and \colorbox{c3}{third best}.}
\begin{tabular}{ll|ccccccccccccc}  
\toprule
Metrics               & Model         & SIREN & MFN   &WIRE   & ReLU  & ReLU+BN & ReLU+LN & ReLU+GN & ReLU+CN & PE    & PE+BN & PE+LN & PE+GN & PE+CN \\
\midrule
\multirow{2}{*}{PSNR} &2$\times$256   & 26.04 & 25.31 & 27.24 & 24.52 & 28.97   & 27.91   & 27.11   & 29.34   & 30.17 & \cellcolor{c2}32.42 & \cellcolor{c3}31.97 & 31.39 & \cellcolor{c1}32.86 \\    
                      &4$\times$256   & 27.89 & 29.74 & 30.24 & 26.11 & 30.62   & 29.71   & 29.58   & 31.09   & 32.12 & \cellcolor{c2}34.56 & \cellcolor{c3}34.21 & 34.11 & \cellcolor{c1}34.97 \\    
\midrule
\multirow{2}{*}{SSIM} &2$\times$256   & 0.708 & 0.623 & 0.722 & 0.781 & 0.888   & 0.857   & 0.825   & 0.899   & 0.902 & \cellcolor{c2}0.914 & \cellcolor{c3}0.902 & 0.858 & \cellcolor{c1}0.918 \\    
                      &4$\times$256   & 0.764 & 0.820 & 0.837 & 0.868 & 0.914   & 0.898   & 0.891   & 0.920   & 0.929 & \cellcolor{c2}0.955 & \cellcolor{c3}0.937 & 0.932 & \cellcolor{c1}0.961 \\
\bottomrule
\end{tabular}
\label{mri-results}
\end{table*}

\begin{figure*}[!t]
  \centering
  \includegraphics[width=\linewidth]{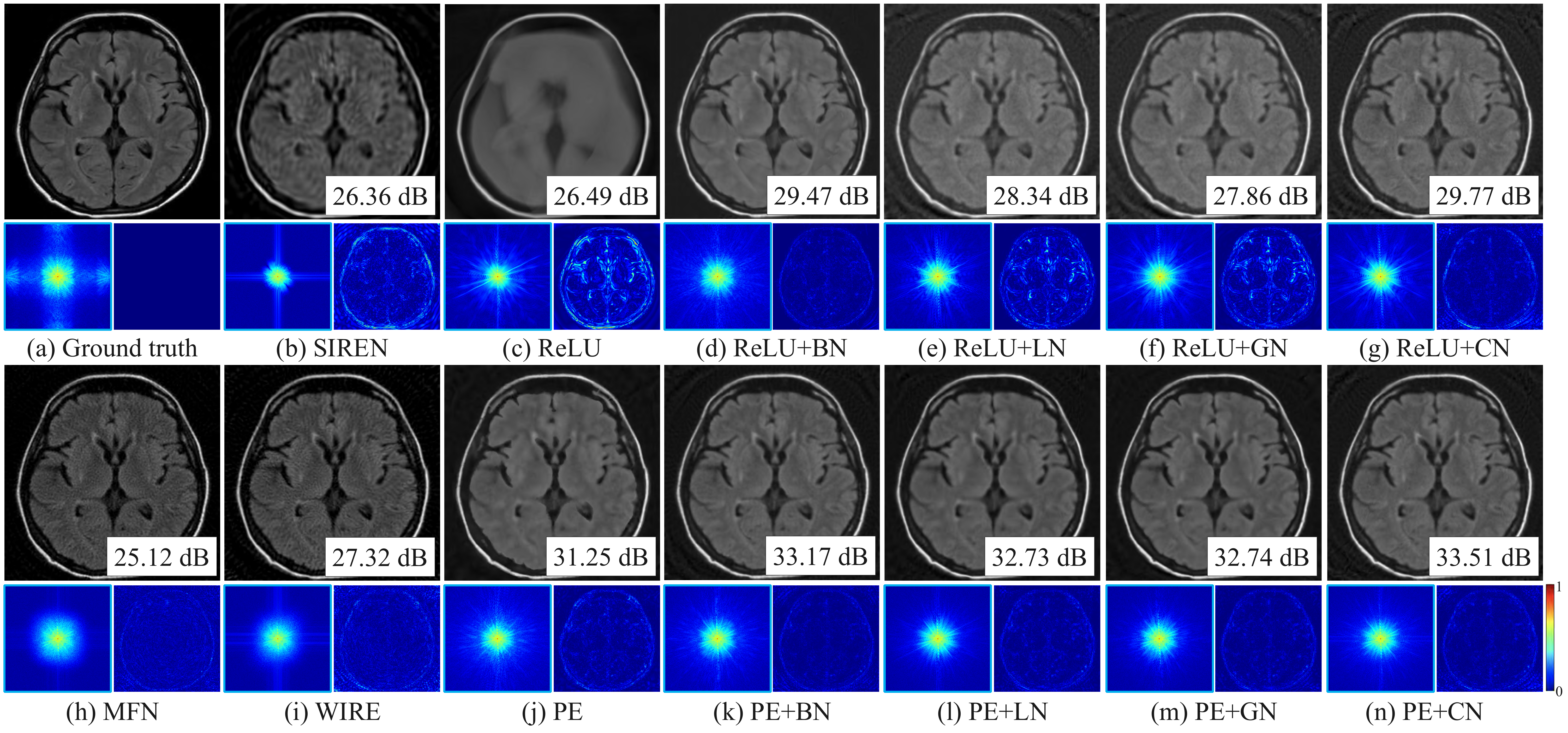}
  \caption{MRI reconstruction from various coordinate networks.}
  \label{mri}
\end{figure*}

\subsection{3D Magnetic Resonance Imaging}
\textbf{Configurations.}
For the 3D MRI task, we observe measurements which are the Fourier transform coefficients of the atomic response to radio waves under a magnetic field. We train a network $f: \mathbb{R}^3\mapsto\mathbb{R}^1$ that takes in 3D voxel coordinates and predicts the corresponding intensity at each location with an indirect supervision. 
We conduct experiments on the ATLAS brain dataset~\cite{tancik2020fourier}, each sample has a volume resolution of $96^3$.
We use networks of two architecture configurations, that is, [2$\times$256] and [4$\times$256]. 
In the experiment, all the models are trained for 1,000 iterations using Adam optimizer with an initial learning rate $2e\!-\!3$.

\textbf{Results.}
Tab.~\ref{mri-results} exhibits the experimental results of various methods evaluated by PSNR.
As can be observed, normalizations considerably improve the performance of simple ReLU, up to $4.51$dB, $3.60$dB, $3.47$dB, and $4.98$dB higher PSNR when equipping the ReLU network of [4$\times$256] with BN, LN, GN, and CN, respectively. 
ReLU+BN and ReLU+CN also significantly surpass the methods with frequency-related activation functions (\textit{i.e.}, SIREN, MFN and WIRE), and ReLU+LN and ReLU+GN are comparable with these methods.
In addition, normalizations also boost the performance of PE, up to $2.44$dB, $2.09$dB, $1.99$dB, and $2.85$dB higher PSNR when equipping the PE network of [4$\times$256] with BN, LN, GN, and CN, respectively.
Moreover, PE+CN still consistently achieves the highest PSNR and SSIM among all the thirteen methods for both network configurations.

\textbf{Visualization.}
Fig.~\ref{mri} visualizes one MRI slice reconstructed by these methods with the network of [2$\times$256].
For the methods with frequency-related activation functions, the reconstructed slices are full of unpredictable textures, indicating the over-fitting to the noise. While networks with normalizations have distinct details and clear background, indicating the superior representational capacity for the target signal and robustness to the noise.

\section{Applications}
Apart from the validation of spectral-bias alleviation by applying normalization techniques to four representation and simple optimization tasks mentioned above, in this section, we will focus on two more complex tasks, \textit{i.e.}, the novel view synthesis and multi-view stereo reconstruction, and demonstrate how the proposed cross normalization further improves the state-of-the-arts.

\begin{table*}[!t]
\centering
\setlength\tabcolsep{11pt}
\caption{Results of different coordinate networks on 5D static novel view synthesis task, measured in PSNR and SSIM.  We color code each cell as \colorbox{c1}{best}, \colorbox{c2}{second best}, and \colorbox{c3}{third best}.}
\begin{tabularx}{0.98\textwidth}{@{}ll|cccccccc|c@{}} 
\toprule
Metrics               & Scene        & Chair & Drums & Ficus & Hotdog& Lego  & Materials & Mic   & Ship  & Average \\
\midrule
\multirow{7}{*}{PSNR} & NeRF         & 33.90 & 25.61 & 28.94 & 36.73 & 31.54 & 29.39     & 33.08 & 29.17 & 31.04   \\    
                      & Plenoxel     & 33.74 & 25.44 & 29.59 & 36.03 & 31.71 & 29.18     & 32.95 & 29.25 & 30.98   \\
                      & DVGO         & 36.68 & 26.31 & \cellcolor{c3}32.69 & \cellcolor{c3}38.38 & 34.73 & \cellcolor{c3}30.31     & 35.32 & \cellcolor{c3}31.26 & 33.21   \\
                      & DINER        & \cellcolor{c3}36.89 & \cellcolor{c3}26.32 & 32.19 & 37.90 & 35.13 & 30.16     & 35.33 & 30.97 & 33.11   \\
                      & Instant NGP  & 36.88 & 26.24 & \cellcolor{c1}33.03 & 38.24 & \cellcolor{c1}35.78 & 29.80     & \cellcolor{c3}36.48 & \cellcolor{c2}31.75 & \cellcolor{c2}33.53   \\
                      & MipNeRF      & \cellcolor{c2}36.93 & \cellcolor{c2}26.33 & \cellcolor{c2}32.80 & \cellcolor{c2}38.46 & \cellcolor{c3}35.25 & \cellcolor{c2}31.52     & \cellcolor{c2}36.62 & 30.78 & \cellcolor{c2}33.58   \\
                      & Ours         & \cellcolor{c1}37.11 & \cellcolor{c1}26.54 & 32.52 & \cellcolor{c1}38.70 & \cellcolor{c2}35.56 & \cellcolor{c1}31.63     & \cellcolor{c1}37.20 & \cellcolor{c1}32.00  & \cellcolor{c1}33.91   \\
\midrule
\multirow{7}{*}{SSIM} & NeRF         & 0.975 & 0.930 & 0.963 & \cellcolor{c3}0.978 & 0.964 & 0.957     & 0.978 & 0.875 & 0.953   \\  
                      & Plenoxel     & 0.973 & 0.932 & 0.969 & 0.977 & 0.967 & 0.958     & 0.977 & 0.878 & 0.954   \\
                      & DVGO         & \cellcolor{c2}0.987 & 0.937 & \cellcolor{c3}0.982 & \cellcolor{c1}0.986 & \cellcolor{c3}0.983 & 0.962     & 0.987 & \cellcolor{c3}0.909 & \cellcolor{c2}0.966   \\
                      & DINER        & \cellcolor{c1}0.988 & 0.939 & 0.980 & \cellcolor{c2}0.984 & \cellcolor{c2}0.984 & \cellcolor{c3}0.963     & 0.987 & 0.908 & \cellcolor{c2}0.966   \\
                      & Instant NGP  & \cellcolor{c3}0.985 & \cellcolor{c3}0.940 & \cellcolor{c3}0.982 & 0.974 & \cellcolor{c1}0.986 & 0.956     & \cellcolor{c3}0.988 & \cellcolor{c2}0.912 & \cellcolor{c3}0.964   \\
                      & MipNeRF      & \cellcolor{c1}0.988 & \cellcolor{c2}0.941 & \cellcolor{c1}0.984 & \cellcolor{c1}0.986 & 0.982 & \cellcolor{c2}0.971     & \cellcolor{c2}0.991 & 0.886 & \cellcolor{c2}0.966   \\
                      & Ours         & \cellcolor{c1}0.988 & \cellcolor{c1}0.943 & \cellcolor{c2}0.983 & \cellcolor{c1}0.986 & \cellcolor{c3}0.983 & \cellcolor{c1}0.972     & \cellcolor{c1}0.992 & \cellcolor{c1}0.916  & \cellcolor{c1}0.970   \\
\bottomrule
\end{tabularx}
\label{nerf-results}
\end{table*}

\begin{figure*}[!t]
  \centering
  \includegraphics[width=\linewidth]{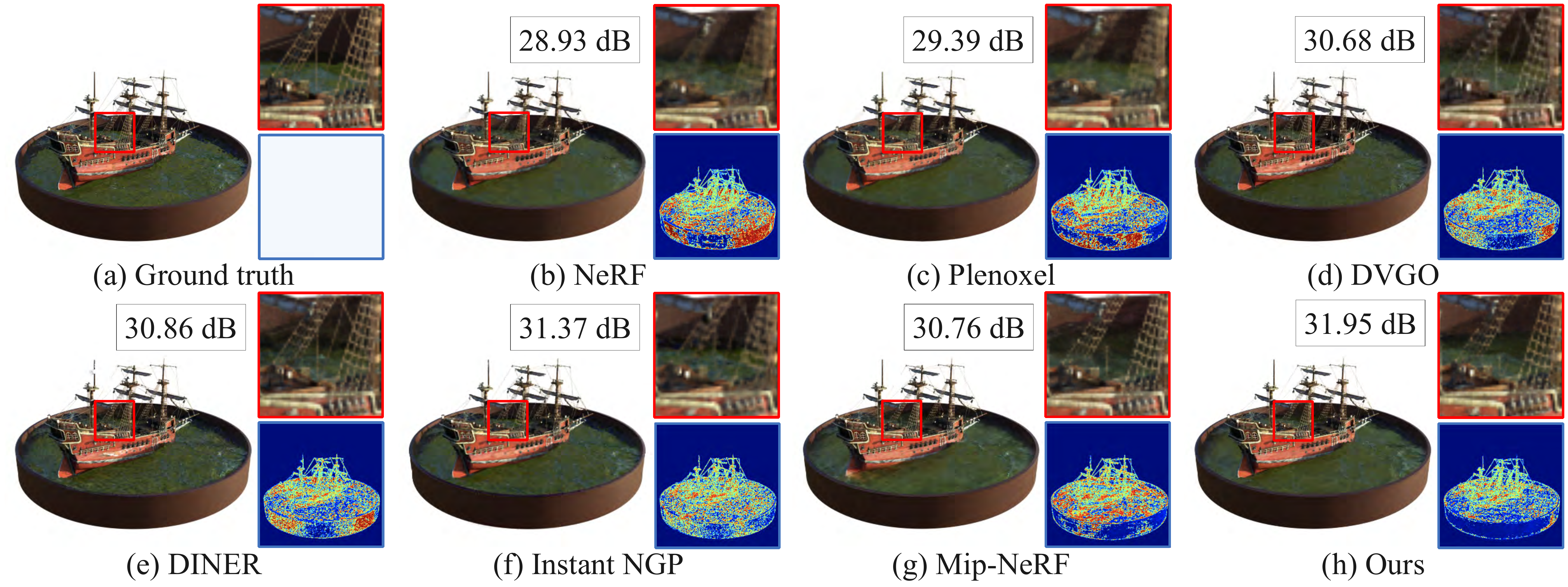}
  \caption{Neural radiance fields optimization with various methods. The corresponding error maps are also visualized.}
  \label{fig:nerf}
\end{figure*}

\begin{table*}[t]
\centering
\setlength\tabcolsep{3pt}
\scriptsize
\caption{Results of different coordinate networks on 3D MVS task. The results are measured in chamfer distance (the lower, the better). We color code each cell as \colorbox{c1}{best}, \colorbox{c2}{second best}, and \colorbox{c3}{third best}.} 
\begin{tabular}{l|ccccccccccccccc|c} 
\toprule
Method     & Scan24 & Scan37 & Scan40 & Scan55 & Scan63 & Scan 65 & Scan69 & Scan83 & Scan97 & Scan105& Scan106& Scan110& Scan114& Scan118& Scan122& Average\\
\midrule
Instant NGP& 1.68   & 1.93   & 1.57   & 1.16   & 2.00   & 1.56    & 1.81   & 2.33   & 2.16   & 1.88   & 1.76   & 2.32   & 1.86   & 1.80   & 1.72   & 1.84 \\
NeRF       & 1.42   & 1.61   & 1.67   & 0.80   & 1.86   & 1.29    & 1.20   & 1.60   & 1.65   & 1.11   & 0.95   & 2.10   & 0.98   & 1.30   & 0.92   & 1.36 \\ 
IDR        & 1.63   & 1.87   & 0.63   & \cellcolor{c3}0.48   & \cellcolor{c2}1.04   & 0.79    & 0.77   & \cellcolor{c2}1.33   & 1.16   & \cellcolor{c3}0.76   & 0.67   & \cellcolor{c1}0.90   & 0.42   & \cellcolor{c3}0.51   & \cellcolor{c3}0.53   & 0.90 \\
VOLSDF     & 1.14   & \cellcolor{c3}1.26   & 0.81   & 0.49   & 1.25   & \cellcolor{c3}0.70    & \cellcolor{c3}0.72   & \cellcolor{c1}1.29   & 1.18   & \cellcolor{c1}0.70   & \cellcolor{c3}0.66   & \cellcolor{c2}1.08   & 0.42   & 0.61   & 0.55   & 0.86 \\
HF-NeuS    & \cellcolor{c3}1.11   & 1.28   & \cellcolor{c3}0.61   & \cellcolor{c2}0.47   & \cellcolor{c1}0.97   & \cellcolor{c2}0.68    & \cellcolor{c2}0.62   & \cellcolor{c3}1.34   & \cellcolor{c1}0.91   & \cellcolor{c2}0.73   & \cellcolor{c2}0.53   & 1.82   & \cellcolor{c3}0.38   & 0.54   & \cellcolor{c2}0.51   & \cellcolor{c3}0.83 \\ 
NeuS       & \cellcolor{c2}0.93   & \cellcolor{c2}1.08   & \cellcolor{c2}0.56   & \cellcolor{c1}0.37   & \cellcolor{c3}1.13   & 0.72    & \cellcolor{c1}0.61   & 1.46   & \cellcolor{c3}1.00   & 0.82   & \cellcolor{c1}0.52   & 1.48   & \cellcolor{c2}0.36   & \cellcolor{c2}0.45   & \cellcolor{c1}0.49   & \cellcolor{c2}0.80 \\
Ours       & \cellcolor{c1}0.91   & \cellcolor{c1}0.86   & \cellcolor{c1}0.55   & \cellcolor{c1}0.37   & 1.16   & \cellcolor{c1}0.67    & \cellcolor{c1}0.61   & 1.43   & \cellcolor{c2}0.93   & 0.78   & \cellcolor{c2}0.53   & \cellcolor{c3}1.22   & \cellcolor{c1}0.35   & \cellcolor{c1}0.42   & \cellcolor{c1}0.49   & \cellcolor{c1}0.75 \\ 
\bottomrule
\end{tabular}
\label{mvs-results}
\end{table*}

\begin{figure}[t]
  \centering
  \includegraphics[width=\linewidth]{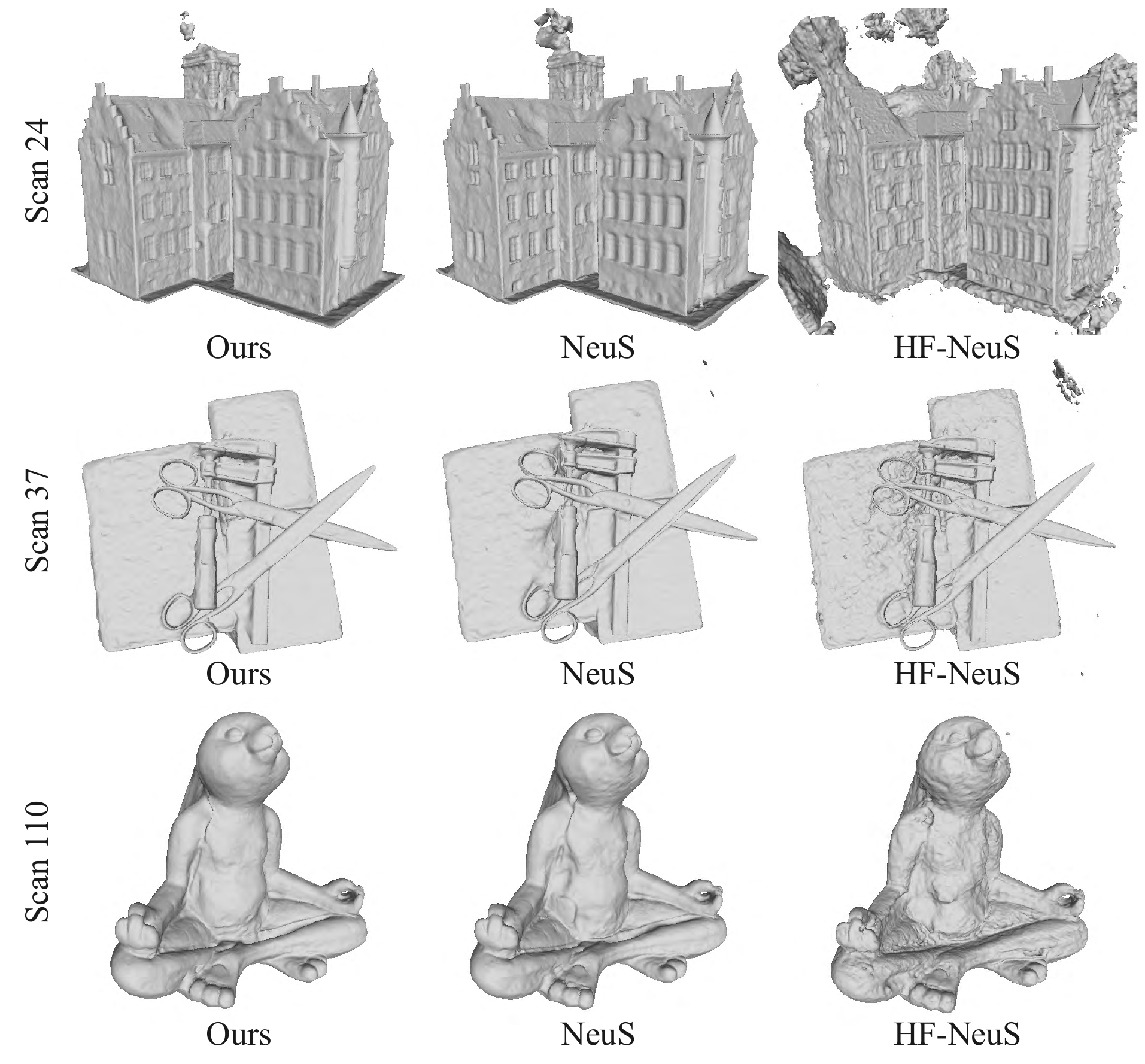}
  \caption{Multi-view stereo reconstruction with various methods.}
  \label{fig:neus}
\end{figure}

\subsection{5D Novel View Synthesis}
\textbf{Configurations.} 
In this section, we demonstrate the improvements of CN on novel view synthesis using the neural radiance fields (NeRF)~\cite{mildenhall2021nerf}. NeRF models the 3D world as a 5D radiance fields using coordinate networks, where the input contains the 3D position and 2D viewing direction of a point and the output attributes include the RGB color and point density, namely, it learns a 5D mapping function $f: \mathbb{R}^5\mapsto\mathbb{R}^4$. Then the color of each pixel is calculated by querying the above attributes along the ray defined by the pixel position and camera's parameters and applying the volume rendering techniques~\cite{max1995optical}. Finally, the radiance field is optimized by supervising rendered color with the ground truth one. Once the radiance field is convergent, the image from any view could be synthesized by following the second step mentioned above. 

In our experiment, six baselines are compared, namely, NeRF~\cite{mildenhall2021nerf}, Plenoxel~\cite{fridovich2022plenoxels}, DVGO~\cite{sun2022direct}, DINER~\cite{zhu2023disorder}, Instant NGP~\cite{muller2022instant}, and the currently SOTA method MipNeRF~\cite{mildenhall2021nerf}. We follow all authors' default configurations.
To better verify the effectiveness, we apply CN to the most powerful baseline MipNeRF, \textit{i.e.}, the term `Ours' in Tab.~\ref{nerf-results} and Fig.~\ref{fig:nerf}.

\textbf{Results and visualization.} Tab.~\ref{nerf-results} lists the quantitative comparisons on the down-scaled Blender dataset~\cite{mildenhall2021nerf} (resolution of $400\times400$). 
Compared with original results of MipNeRF, applying CN notably improves the average PSNR up to $0.33$dB. 
In addition, CN-enhanced MipNeRF achieves the best results among all the methods, delivering the new state-of-the-art. 
Fig.~\ref{fig:nerf} qualitatively compares the reconstructed details on the `Ship' dataset. The corresponding error maps are also visualized. It is noticed that additional CN significantly reduces the errors, and more details are reconstructed such as the zoomed-in cable wind rope.

\subsection{3D Multi-view Stereo Reconstruction}
\textbf{Configurations.}
In this section, we demonstrate the improvements of CN on multi-view stereo reconstruction using the neural implicit surfaces (NeuS)~\cite{wang2021neus}. 
NeuS reconstructs the surfaces of objects and scenes with high fidelity from multi-view 2D images, which extends the basic NeRF formulation by integrating an signed distance function (SDF) into volume rendering.
With NeuS, the scene of an object to be reconstructed is represented by two functions, namely, $f: \mathbb{R}^3\mapsto\mathbb{R}$ that maps a spatial position $x\in\mathbb{R}^3$ to its signed distance to the object, and $c: \mathbb{R}^3\times\mathbb{S}^2\mapsto\mathbb{R}^3$ that encodes the RGD color associated with a point $x\in\mathbb{R}^3$ and a viewing direction $v\in\mathbb{S}^2$. Both functions are encoded by coordinate networks. The surface $\mathcal{S}$ of the object is represented by the zero-level set of its SDF $f(x)$, namely, $\mathcal{S}=\{x\in\mathbb{R}^3|f(x)=0\}$.
Finally, NeuS employs specific volume rendering with the aid of probability density function $\phi_s(f(x))$~\cite{wang2021neus} to recover the underlying SDF, which is optimized by minimizing the photometric loss between the renderings and ground-truth images.

In our experiment, six baselines are compared, namely, Instant NGP~\cite{muller2022instant}, NeRF~\cite{mildenhall2021nerf}, IDR~\cite{yariv2020multiview}, VOLSDF~\cite{yariv2021volume}, HF-NeuS~\cite{wang2022hf}, and the NeuS\cite{wang2021neus}. We follow all authors' default configurations.
To better verify the effectiveness, we apply CN to the most powerful baseline NeuS, \textit{i.e.}, the term `Ours' in Tab.~\ref{mvs-results} and Fig.~\ref{fig:neus}.

Following the setting of previous work~\cite{yariv2020multiview,wang2022hf,wang2021neus}, we conduct our experiments on the 15 scenes from the DTU dataset~\cite{jensen2014large}, with a wide variety of materials, appearance and geometry, including challenging cases for reconstruction algorithms, such as non-Lambertian surfaces and thin structures. Each scene contains 49 or 64 images with the image resolution of $1600\times1200$. We adopt the foreground masks provided by IDR~\cite{yariv2020multiview} for these scenes.

\textbf{Results and visualization.}
Tab.~\ref{mvs-results} lists the quantitative comparisons on the DTU dataset, evaluated in chamfer distance. 
Compared with original results of NeuS, applying CN notably improves the average chamfer distance up to $6.25\%$. 
In addition, CN-enhanced NeuS also achieves the best results among all the methods, delivering the new state-of-the-art. 
Fig.~\ref{fig:neus} qualitatively compares the reconstructed details on three different scenes. 

\section{Conclusion}
In this paper, we leverage the mean field theory to prove that classical normalization techniques (batch normalization and layer normalization) can optimize the NTK's eigenvalues distribution, enabling the coordinate networks overcome the spectral bias, namely, better suited for learning the high-frequency components. 
Based on the observation of the theoretical analysis, we propose two new normalization methods, namely, global normalization and cross normalization.
We also empirically show that adding these four normalizations transit the NTK's eigenvalues distribution towards larger values for both standard MLP and position encoding-enhanced version, which are also experimentally validated to be capable of accelerating the convergence of high-frequency components significantly.
Extensive experiments substantiate that norm-based networks significantly surpass existing methods on various representation and inverse optimization tasks, obtaining the new state-of-the-art performance.





\ifCLASSOPTIONcaptionsoff
  \newpage
\fi



%



{\small
\input{main.bbl}

}
\end{document}